\documentclass{article} 
\usepackage[final]{colm2025_conference}

\usepackage{microtype}
\usepackage{hyperref}
\usepackage{url}
\usepackage{booktabs}

\usepackage{lineno}

\definecolor{darkblue}{rgb}{0, 0, 0.5}
\hypersetup{colorlinks=true, citecolor=darkblue, linkcolor=darkblue, urlcolor=darkblue}

\usepackage{tcolorbox}
\usepackage{amsmath}
\usepackage{multirow}

\usepackage{hhline}
\definecolor{bg}{rgb}{0.9, 0.9, 0.9}
\usepackage{bibentry}

\usepackage{fancyvrb,fvextra}

\usepackage{inconsolata}
\usepackage{graphicx}
\usepackage{soul}
\usepackage{multicol}
\usepackage{multirow}
\usepackage{amssymb}

\usepackage{microtype}

\usepackage{multirow}
\usepackage{xcolor,colortbl}
\usepackage{amsmath,array,graphicx}
\usepackage{hhline}
\usepackage{rotating}
\usepackage{graphicx} 

\usepackage{wrapfig}
\usepackage{caption}
\usepackage{enumitem}

\usepackage[T1]{fontenc}

\usepackage[normalem]{ulem}
\useunder{\uline}{\ul}{}

\title{mSCoRe: a $M$ultilingual and Scalable Benchmark for $S$kill-based $Co$mmonsense $Re$asoning}

\author{Nghia Trung Ngo\textsuperscript{\rm 1}, Franck Dernoncourt\textsuperscript{\rm 2} and Thien Huu Nguyen\textsuperscript{\rm 1} \\
\textsuperscript{\rm 1} Department of Computer Science, University of Oregon, Eugene, OR, USA \\ 
\textsuperscript{\rm 2} Adobe Research, USA\\
  \texttt{\{nghian@,thien@cs\}.uoregon.edu}, \texttt{franck.dernoncourt@adobe.com} \\
}

\begin{document}

\ifcolmsubmission
\linenumbers
\fi

\maketitle

\begin{abstract}
Recent advancements in reasoning-reinforced Large Language Models (LLMs) have shown remarkable capabilities in complex reasoning tasks.
However, the mechanism underlying their utilization of different human reasoning skills remains poorly investigated, especially for multilingual commonsense reasoning that involves everyday knowledge across different languages and cultures.
To address this gap, we propose a \textbf{M}ultilingual and Scalable Benchmark for \textbf{S}kill-based \textbf{Co}mmonsense \textbf{Re}asoning (\textbf{mSCoRe}).
Our benchmark incorporates three key components that are designed to systematically evaluate LLM's reasoning capabilities, including: 
(1) a novel taxonomy of reasoning skills that enables fine-grained analysis of models' reasoning processes,  
(2) a robust data synthesis pipeline tailored specifically for commonsense reasoning evaluation, and
(3) a complexity scaling framework allowing task difficulty to scale dynamically alongside future improvements in LLM abilities.
Extensive experiments on eights state-of-the-art LLMs of varying sizes and training approaches demonstrate that \textbf{mSCoRe} remains significantly challenging for current models, particularly at higher complexity levels.
Our results reveal the limitations of such reasoning-reinforced models when confronted with nuanced multilingual general and cultural commonsense.
We further provide detailed analysis on the models' reasoning processes, suggesting future directions for improving multilingual commonsense reasoning capabilities.
\end{abstract}

\vspace{-3mm}
\section{Introduction}
\vspace{-2mm}
Commonsense reasoning enables a person to navigate everyday situations, make logical inferences, and understand implicit information in our environment. 
While this ability comes naturally to humans, it has proven to be one of the most challenging capabilities to replicate in current language models \citep{survey_ernest:23}.
Recent advancements in Large Reasoning Models (LRMs), such as OpenAI's o1 series \citep{o1:24}, and open-source models like DeepSeek R1 \citep{r1:25}, have shown promising results across various complex reasoning tasks, including mathematics, coding, and logical inference \citep{bbeh:25}.
However, relatively little attention has been devoted to systematically analyzing and understanding these models' commonsense reasoning capabilities, especially in multilingual settings which involve common knowledge across diverse languages and cultural contexts \citep{what-cms:24}.

Several benchmarks have been proposed to assess commonsense reasoning abilities of language models. 
CommonsenseQA (CSQA) \citep{csqa:19} evaluates general commonsense knowledge through multiple-choice questions derived from ConceptNet. COPA \citep{copa:11} focuses on causal relationships between everyday events, while SocialIQA \citep{socialiqa:19} evaluates social commonsense understanding.
More recently, comprehensive benchmarks like MMLU \citep{mmlu:21} and Big-Bench Hard \citep{bbh:23} aim to evaluate model's generalization capabilities across diverse commonsense tasks.
However, these benchmarks have significant limitations in three key areas.
First, they often focus on a single high-resourced language such as English \citep{csqa:19} or Chinese \citep{chinese:24}. Multilingual extensions such as X-COPA \citep{xcopa:20} and X-CSQA \citep{xcsqa:21} primarily rely on translation of existing datasets, thus limiting their ability to capture culturally specific nuances.
Second, although recent efforts like mCSQA \citep{mcsqa:24} leverage generative multilingual language models for a more comprehensive and robust dataset creation process, they still lack a systematic way to scale task difficulty, which is crucial for assessing the rapid evolving capabilities of LLMs.
Finally, current benchmarks are unable to provide fine-grained analysis and classification of the reasoning steps used by LLMs, which would provide deeper insights into their operations.

To address these limitations, we introduce \textbf{M}ultilingual and \textbf{S}calable Benchmark for \textbf{S}kill-based \textbf{Co}mmonsense \textbf{Re}asoning (\textbf{mSCoRe}), a novel benchmark designed explicitly to provide a comprehensive evaluation of LLMs' commonsense reasoning capabilities across multiple languages and cultural contexts. Specifically, our benchmark offers three notable advantages:

\setlist{nolistsep}
\begin{enumerate}[leftmargin=*,noitemsep]
\item \textbf{Comprehensive Coverage:} \textbf{mSCoRe} covers both \texttt{general} commonsense knowledge from across languages including English, German, French, Chinese, and Japanese, and diverse cultural \texttt{social} commonsense knowledge.

\item \textbf{Skill-based Analysis:} \textbf{mSCoRe} introduces a novel approach to reasoning analysis through the classification of each atomic reasoning step, allowing for more precise analysis of model's reasoning process.

\item \textbf{Scalability:} \textbf{mSCoRe} employs techniques such as context expansion, option adjustment, and commonsense implicitation to progressively increase question complexity while preserving commonsense answer semantics, effectively scaling task difficulty.
\end{enumerate}

Our contributions can be summarized as follows:
\setlist{nolistsep}
\begin{itemize}[leftmargin=*,noitemsep]
    \item We introduce \textbf{mSCoRe}, a novel scalable benchmark for evaluating multilingual general and cultural commonsense reasoning with fine-grained skill-based analysis.
    \item Using \textbf{mSCoRe}, we extensively evaluate eights state-of-the-art LLMs, including both commercial and open-source models, across diverse reasoning conditions.
    \item Our analysis provides insights into how model scale, training techniques, and reasoning skill types impact performance, suggesting future directions for improving commonsense reasoning capability of LLMs.
\end{itemize}

\vspace{-1mm}
\section{Related works}
\vspace{-2mm}

\textbf{Large Reasoning Models: }
Recent advancements in Large Language Models (LLMs) have demonstrated remarkable capabilities in various complex problem-solving tasks. 
Reasoning-reinforced models like OpenAI o1 \citep{o1:24}, Macro-o1 \citep{macro:24} and DeepSeek-R1 \citep{r1:25} have shown superior performance in mathematics and coding, effectively simulating human-like analytical thinking and enhancing multi-step reasoning \citep{front-math:24, r1-code:24}. 
These models employ multiple methods to enhance reasoning capabilities. 
In particular, Chain-of-thought prompting \citep{cot:23} has emerged as a powerful technique that encourages step-by-step reasoning, significantly improving performance on complex tasks.
Building upon this foundation, various Chain-of-X approaches have been proposed to further enhance model reasoning capability \citep{tree:23, verify:24, graph:24, boost:24}. 
Recent techniques such as test-time scaling and reinforcement learning have also contributed to improving the reasoning abilities of LLMs \citep{scale-optim:24, llm-post:25, rein-inf:25}.
While these methods enhance the overall structure of the reasoning path, they generally pay little attention to the categorization of each reasoning step.
\textbf{mSCoRe} proposes a more fine-grained approach in which each step is atomic and labeled according to a reasoning skill, facilitating a deeper and systematic evaluation of model's reasoning process.

\begin{figure*}[ht!]
\addtolength{\abovecaptionskip}{-2mm}
\addtolength{\belowcaptionskip}{-5mm}
\begin{center} 
\includegraphics[width=0.97\textwidth]{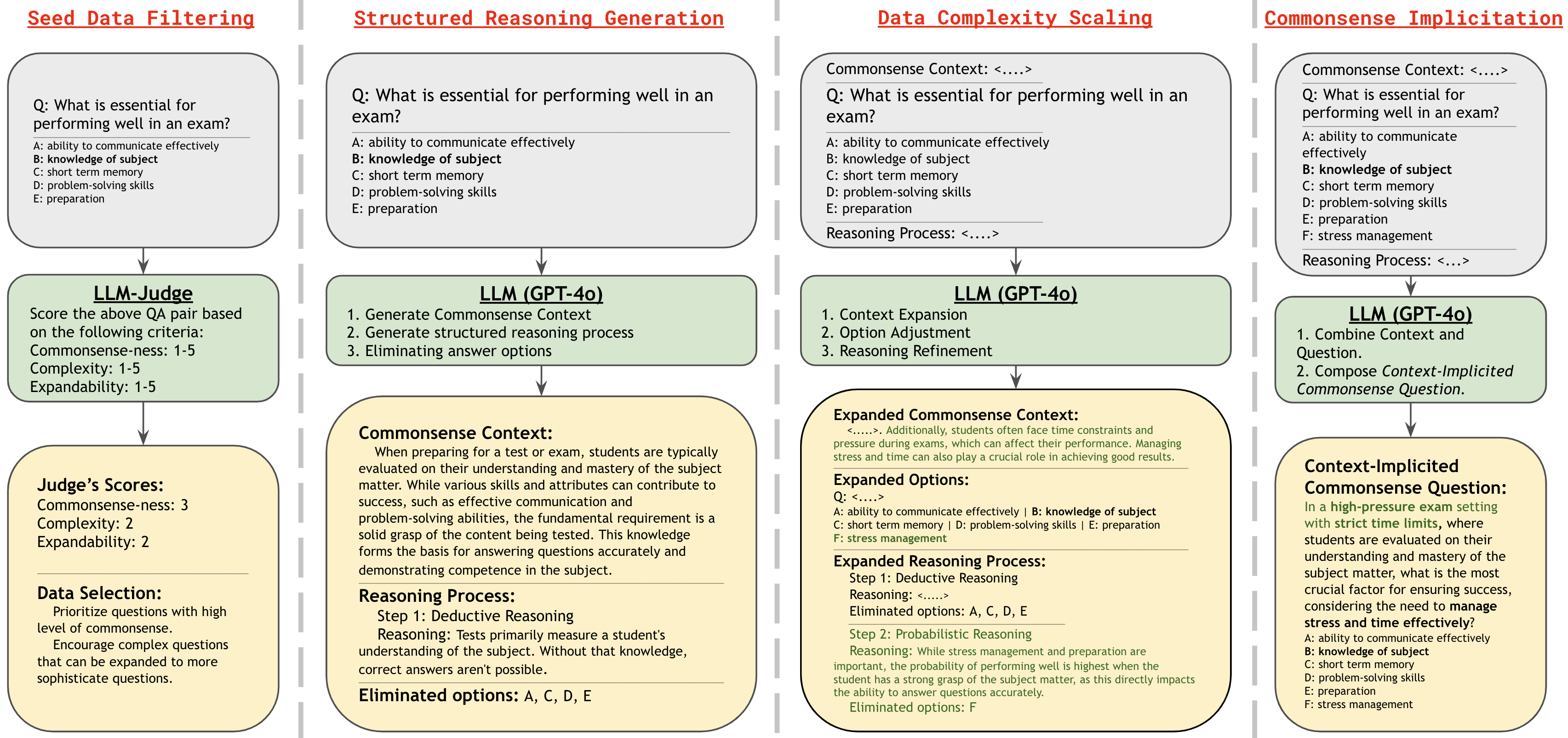}
\captionsetup{font=footnotesize}
\caption{\footnotesize \textbf{Data Generation Process. }
The four-step data creation pipeline of \textbf{mSCoRe}.
Each step builds upon the previous one to create progressively more challenging reasoning tasks while maintaining the underlying reasoning skills being evaluated.}
\label{fig:data_gen}
\end{center}
\end{figure*}

\textbf{Commonsense Reasoning Benchmarks: }
Despite significant advances in evaluating mathematical and scientific reasoning capabilities of LLMs \citep{math-word:21, front-math:24, olym:24, phys:25}, commonsense reasoning benchmarks have received comparatively less recent attention. 
Early datasets such as CommonsenseQA (CSQA) \citep{csqa:19}, COPA \citep{copa:11}, and SocialIQA \citep{socialiqa:19} primarily target English-language commonsense knowledge, focusing respectively on general factual knowledge, causal relationships, and social interactions.
Recent comprehensive benchmarks like MMLU \citep{mmlu:21} and Big-Bench Hard \citep{bbh:23} evaluate the generalization abilities of LLMs across diverse commonsense reasoning tasks.
Multilingual extensions like X-CSQA \citep{xcsqa:21} and X-COPA \citep{xcopa:20} expand the evaluation beyond English by translating existing datasets into multiple languages.
More recent approaches such as mCSQA \citep{mcsqa:24} leverage LLM to assist more closely in the data synthesis process.
However, these benchmarks are still limited in cultural social commonsense that involves everyday interactions among different cultures.
There has been increasing efforts on cultural knowledge bases to develop cultural-aware LLMs.
In particular, CulturePark \citep{culpark:24} introduces a novel multi-agent communication framework powered by LLMs to simulate cross-cultural human interactions, whereas CultureBank \citep{culbank:24} aggregates real-world social interactions from platforms like TikTok and Reddit, structuring annotations around cultural topics.
\textbf{mSCoRe} builds upon mCSQA and CultureBank to comprehensively cover both \textit{general} and \textit{social} aspects of commonsense reasoning across multiple languages and cultures.

\vspace{-1mm}
\section{Benchmark Creation}
\vspace{-2mm}


\begin{wrapfigure}{R}{0.35\textwidth}
\vspace{-8mm}
\addtolength{\abovecaptionskip}{-3mm}
\addtolength{\belowcaptionskip}{-2mm}
\fcolorbox{bg}{bg}{
\scriptsize
\begin{minipage}{0.32\textwidth}
\textbf{Atomic Reasoning Step} \  an indivisible unit of reasoning that predominantly utilizes one reasoning skill. It is a single, coherent thought process that cannot be broken down into smaller steps without losing its meaning. 
An optimal reasoning path (for multiple-choice QA task) uses a minimum number of atomic steps necessary, ensuring that each step is non-redundant and contributes to narrowing down the possible options by eliminating one or more answer choices.
\end{minipage}
}
\captionsetup{font=scriptsize}
\caption{\scriptsize Atomic Reasoning Step definition.}
\label{fig:atomic_def}
\end{wrapfigure}

\subsection{Commonsense Reasoning} \label{3.1-taxonomy}
Commonsense reasoning involves making inferences about unstated aspects of a scenario using implicit world knowledge – a capability ingrained in human behavior but still challenging for current LLMs.
Unlike formal reasoning domains such as mathematics or logic, where rules are explicitly defined and conclusions follow determinate paths, commonsense reasoning requires access to a vast reservoir of implicit knowledge and the ability to apply this knowledge flexibly across diverse situations.
Furthermore, there can be multiple reasoning paths that can lead to the correct answer, especially for commonsense questions.
However, previous evaluations still primarily focus on answer accuracy \citep{bbh:23, mcsqa:24}, providing limited insight into how LLMs construct their reasoning pathways.

To address this limitation, we propose to investigate deeply into model's reasoning process by introducing the concept of ``\textit{atomic reasoning steps}" (Fig. \ref{fig:atomic_def}) as the foundational unit of analysis.
Our framework aims to analyze the optimal path utilized by the LLM, defined as the path requiring the minimum number of atomic reasoning steps while maintaining logical coherence.
This approach not only enables systematic evaluation of specific reasoning skills, but also provides a clear framework for analyzing how models construct complex reasoning chains. 
It allows for meaningful comparison of reasoning processes across different models and languages. 
Finally, it facilitates the scaling of question complexity through the requirement of additional of atomic steps. 


\vspace{-1mm}
\subsubsection{Reasoning Skills}
\vspace{-1mm}
We develop a structured taxonomy for classifying each reasoning step, enabling systematic evaluation of how LLMs employ human reasoning skills in commonsense tasks.
While no clear consensus exists on a comprehensive taxonomy of human reasoning skills, existing categorizations typically serve specific purposes. For example, Bloom's Taxonomy \citep{bloom:10} provides a hierarchical framework categorizing educational goals into three domains: cognitive (knowledge-based), affective (emotion-based), and psychomotor (action-based). Similarly, Fleishman's taxonomy \citep{fleishman:86} identifies 52 distinct human abilities across cognitive, perceptual, psychomotor, and physical domains, primarily to facilitate job design, training, and assessment development.
Based on the fundamental characteristics of commonsense knowledge identified by \cite{what-cms:24} with established reasoning skill categorizations from \cite{wiki:cms}, we propose a taxonomy comprising three major categories:

\setlist{nolistsep}
\begin{itemize}[leftmargin=*,noitemsep]
\item \textbf{Logical Reasoning}
encompasses forms of reasoning that involve structured processes to derive conclusions from given information. This category includes methodologies like \textbf{deductive}, \textbf{inductive}, and \textbf{abductive} reasoning, which are foundational in scientific and analytical disciplines to ensure conclusions are logically sound.

\item \textbf{Contextual Reasoning}
includes skills used to understand relationships, contexts, and dynamics between elements. This category covers various types of reasoning such as \textbf{analogical}, \textbf{counterfactual}, \textbf{probabilistic}, \textbf{temporal}, and \textbf{spatial}, used to evaluate scenarios, predict outcomes, and solve problems across different contexts.

\item \textbf{Social and Ethical Reasoning}
involves skills focused on understanding social interactions and evaluating ethical principles. This category includes \textbf{social} and \textbf{moral} reasoning, essential for interpreting behaviors, navigating complex social environments, and making decisions based on ethical considerations.
\end{itemize}

Detailed descriptions and examples of each reasoning skill are provided in Table \ref{tab:skill_detail}. While humans employ additional reasoning skills beyond those presented here, our goal is to establish a concise yet comprehensive reasoning taxonomy that maximizes coverage of human reasoning capabilities for commonsense applications, while minimizing the overlap between categories. 
This reasoning taxonomy will be implemented within our LLM prompts throughout both the data generation and evaluation processes to ensure focus on our considered skills. Each atomic reasoning step will be classified under a single skill from our taxonomy, enabling precise comparison of different reasoning processes.

\begin{table}[]
\addtolength{\abovecaptionskip}{-2mm}
\addtolength{\belowcaptionskip}{-5mm}
\resizebox{0.99\textwidth}{!}
{
\begin{tabular}{lll}
\multicolumn{1}{c|}{Skills} & \multicolumn{1}{c|}{Short Definitions} & \multicolumn{1}{c}{Examples} \\ \hline
\multicolumn{3}{c}{Logical Reasoning} \\ \hline
\multicolumn{1}{l|}{Inductive} & \multicolumn{1}{l|}{Drawing general conclusions from specific observations.} & Most technological innovations eventually benefit society. \\ \hline
\multicolumn{1}{l|}{Deductive} & \multicolumn{1}{l|}{Deriving specific conclusions from general premises.} & All communication tools connect people; social media is a communication tool. \\ \hline
\multicolumn{1}{l|}{Abductive} & \multicolumn{1}{l|}{Forming hypotheses to explain observations.} & Rising depression rates suggest social media affects mental health. \\ \hline \hline
\multicolumn{3}{c}{Contextual Reasoning} \\ \hline
\multicolumn{1}{l|}{Analogical} & \multicolumn{1}{l|}{Drawing parallels between similar situations to infer conclusions.} & Like town squares facilitated discourse, social media creates digital gathering spaces. \\ \hline
\multicolumn{1}{l|}{Counterfactual} & \multicolumn{1}{l|}{Considering alternative scenarios and outcomes that did not happen.} & Without social media, many social movements would lack momentum. \\ \hline
\multicolumn{1}{l|}{Probabilistic} & \multicolumn{1}{l|}{Applying principles of probability to make inferences under uncertainty.} & Users have a very high chance of encountering misinformation weekly. \\ \hline
\multicolumn{1}{l|}{Temporal} & \multicolumn{1}{l|}{Understanding sequences and durations of events.} & Brief moments scrolling accumulate into hours of lost productivity daily. \\ \hline
\multicolumn{1}{l|}{Spatial} & \multicolumn{1}{l|}{Visualizing and manipulating objects in space.} & Platform designs maximize attention capture through strategic layouts. \\ \hline \hline
\multicolumn{3}{c}{Social \& Ethical Reasoning} \\ \hline 
\multicolumn{1}{l|}{Social} & \multicolumn{1}{l|}{Understanding social interactions and norms.} & Like-based validation systems create unhealthy approval-seeking behaviors. \\ \hline
\multicolumn{1}{l|}{Moral} & \multicolumn{1}{l|}{Deciding what is right or wrong based on ethical principles.} & Prioritizing profit over user wellbeing raises ethical concerns.
\end{tabular}
}
\captionsetup{font=scriptsize}
\caption{\scriptsize 
The ten types of reasoning skills across three categories with short definitions and examples for each skill applied to the question \textit{``Is social media good for society?"}. Detailed descriptions and additional examples are provided in Appendix \ref{app:skill}.
}
\label{tab:skill_detail}
\end{table}



\vspace{-2mm}
\subsection{\textbf{mSCoRe}} \label{3.2-mscore}
\vspace{-1mm}

\begin{wrapfigure}{R}{0.5\textwidth}
\captionsetup{font=footnotesize}
\vspace{-5mm}
\addtolength{\abovecaptionskip}{-3mm}
\addtolength{\belowcaptionskip}{-8mm}
\fcolorbox{bg}{bg}{
\scriptsize
\begin{minipage}{0.47\textwidth}
\textbf{Step 1 - Data Filtering:} To limit the cost while maintaining quality and diversity, we sample a small subset from the seed benchmarks. Each sample is scored by a general LLM-judge based on multiple criteria for expansion potential, ensuring that we select instances that will yield meaningful insights when scaled complexity-wise.

\textbf{Step 2 - Reasoning Generation:} Provide a \textit{Commonsense Context} to expand on the given question and a detailed \textit{Reasoning Process} that involves multiple \textit{Reasoning Steps} to arrive at the correct answer. This establishes a gold standard reasoning path for each question.

\textbf{Step 3 - Complexity Scaling:} Modify and expand each question to create more complex variants by expanding its context, modifying the question, adjusting the answer options, and adding additional \textit{Reasoning Steps}. This creates a progression of difficulty levels for each base question.

\textbf{Step 4 - Commonsense Implicitation:} Combine the given \textit{Commonsense Context} with the question to generate a new, concise commonsense question that \textit{implicitly incorporates} the original context. This process aims to evaluate the commonsense reasoning abilities of LLMs by ensuring that the \textit{implicit context} preserves the original reasoning process and maintains the correctness of the answer.

\end{minipage}
}
\caption{\footnotesize Four steps of data generation process.}
\label{fig:gen_step}
\vspace{8mm}

\fcolorbox{bg}{bg}{
\scriptsize
\begin{minipage}{0.47\textwidth}

\textbf{Commonsense-ness:} Does answering the question rely solely on commonsense knowledge accessible to the general population, or does it require formal reasoning and specialized expertise beyond everyday understanding?

\textbf{Complexity:} How difficult is the question to understand and answer? Does it require minimal reasoning or a complex, multi-step thought process to identify the correct answer?

\textbf{Expandability:} To what extent can the question be expanded or elaborated upon to introduce additional complexity or dimensions?

\end{minipage}
}
\caption{\footnotesize Three criteria of data filtering.}
\label{fig:criteria_gen}

\vspace{8mm}

\fcolorbox{bg}{bg}{
\scriptsize
\begin{minipage}{0.47\textwidth}

\textbf{Context Expansion:} Add additional background or situational details to the \textit{Commonsense Context} to increase depth and reasoning requirements to the question.

\textbf{Option Adjustment:} Adjust the existing answer options to align with the new complex question, ensure the correct answer option remains semantically similar to the original. Introduce an additional plausible but incorrect option to increase the complexity of the question that (1) increases the complexity of the question, and (2) requires an additional reasoning step to eliminate.

\textbf{Reasoning Refinement:} Refine the original \textit{Reasoning Process} to fit the new context with an additional reasoning step that eliminates the added incorrect option. 

\end{minipage}
}
\caption{\footnotesize Three sub-steps of Complexity Scaling.}
\label{fig:complexity_step}
\vspace{-3mm}

\end{wrapfigure}

To maintain robust label accuracy, rather than using LLMs to generate a synthetic dataset from scratch, we utilize human-annotated seed datasets and scale up their complexity to create \textbf{mSCoRe}.
In particular, our benchmark consists of multiple-choice commonsense questions, separated into two subsets based on different seed datasets:
(1) \textbf{mSCoRe-G} focuses on general commonsense reasoning, building upon multilingual commonsense questions from mCSQA \citep{mcsqa:24} as a seed dataset. This component evaluates understanding of physical causality, temporal relationships, and basic world dynamics across multiple languages.
(2) \textbf{mSCoRe-S} addresses social commonsense reasoning based on diverse cultural situations from CultureBank \citep{culbank:24}. This component specifically tests understanding of social interactions, cultural norms, and behavioral expectations across different cultural contexts.

The overall data generation process is visualized in Fig. \ref{fig:data_gen}, in which each instance in the seed datasets undergoes a four-step process as illustrated in Fig. \ref{fig:gen_step}.
Through this systematic creation process, \textbf{mSCoRe} provides a comprehensive framework for evaluating and analyzing commonsense reasoning capabilities of LLMs.

\vspace{-1mm}
\subsubsection{\textbf{mSCoRe-G}: General Commonsense}
\vspace{-1mm}

mCSQA (Multilingual CommonsenseQA) extends the CommonsenseQA dataset \citep{csqa:19} to eight languages to evaluate language models' cross-lingual commonsense reasoning capabilities. 
Building upon ConceptNet, each multiple-choice question-answer (QA) pair in mCSQA mostly revolves around general commonsense knowledge (an example is provided in the first step in Figure \ref{fig:data_gen}).
To create \textbf{mSCoRe-G}, we further process each QA pair through the following 4 steps:

\noindent \textbf{1. Seed Data Filtering: } A general LLM-judge evaluates each candidate on three criteria described in Fig. \ref{fig:criteria_gen}: (1) Commonsense-ness, (2) Complexity, and (3) Expandability. The goal is to prioritize questions with a high level of commonsense and complexity, while maintaining flexibility for expansion into more sophisticated questions (full details of the judge model and scoring criteria are provided in Appendix \ref{app:judge}).

\noindent \textbf{2. Structured Reasoning Generation: } For selected question-answer pairs, we employ LLMs to generate relevant commonsense context that helps identify the correct answer. From the tuple (context, question, options-answer), we then generate a structured reasoning process. Each reasoning step in the process consists of three attributes:
(1) \textbf{Reasoning Skill} - the specific skill from our reasoning ontology that is predominantly employed in this step,
(2) \textbf{Reasoning Text} - the model's rationale based on the identified skill, and
(3) \textbf{Eliminated Options} - the list of options eliminated in this step based on the reasoning.

\noindent \textbf{3. Data Complexity Scaling:} From the base (context, question, answer, reasoning process), we implement a procedure to systematically scale up the difficulty level of each question.
The goal is to introduce an additional plausible option at each level that not only increases the complexity of the question, but also requires an additional reasoning step to eliminate. This is achieved through 3 sub-steps, as described in Fig. \ref{fig:complexity_step}.

\noindent \textbf{4. Commonsense Implicitation:} As commonsense knowledge is implicit knowledge about the world that is often unspoken but assumed, this step reduces the context exposed to the LLMs by combining the context and question into a \textit{context-implicit commonsense question}. To answer the modified question, models will have to draw on their internal common knowledge to determine the correct answer, especially when the topic requires more than just logical reasoning.

The whole procedure (Fig. \ref{fig:data_gen}) is repeated to create questions at an increasing level of complexity.
This approach helps mitigate the data leakage \citep{leak:24} and shortcut reasoning \citep{shortcut:23} problems, as observed in our experimental results where performance degrades significantly with each level. Furthermore, the scaled complexity forces LLMs to utilize their reasoning capacity more extensively, enabling deeper investigation into their reasoning process.
Detailed examples with complete prompts are provided in Appendix \ref{app:prompt}.


\vspace{-1mm}
\subsubsection{\textbf{mSCoRe-S}: Social Commonsense}
\vspace{-1mm}
There is still a gap in current commonsense benchmarks in terms of social commonsense knowledge and cultural norms \citep{survey_ernest:23}.
To provide a comprehensive evaluation of LLMs' commonsense reasoning capacity, we propose an additional benchmark \textbf{mSCoRe-S} that revolves around social situations across diverse cultural contexts.
In particular, we utilize CultureBank \citep{culbank:24} as our seed dataset, which is a knowledge base containing real-world social questions sourced from TikTok and Reddit posts. 
Each instance in CultureBank is provided with various descriptors containing details about the cultural group, context, behaviors, and an agreement level indicating how widely accepted that behavior is within the community (Fig. \ref{fig:culbank_example}).

\begin{table}[]
\addtolength{\abovecaptionskip}{-2mm}
\addtolength{\belowcaptionskip}{-5mm}
\resizebox{0.99\textwidth}{!}
{
\begin{tabular}{l|l|l}

\multicolumn{1}{c|}{\textbf{Descriptors}} & \multicolumn{1}{c|}{\textbf{Definitions}} & \multicolumn{1}{c}{\textbf{Examples}} \\ \hline
\textbf{Cultural Topic} & Cultural group - topic - scenario & Japanese culture - Gift Giving - Etiquette and Practices \\ \hline
\textbf{Social Context} & Settings the behavior takes place. & \begin{tabular}[c]{@{}l@{}}During a meeting in Japan, a visiting Western executive\\ wants to express gratitude to their hosts\end{tabular} \\ \hline
\textbf{Actor} & Who exhibit the behavior & Visiting executive \\ \hline
\textbf{Question} & \begin{tabular}[c]{@{}l@{}}The commonsense question regarding \\ the actor's behavior\end{tabular} & \begin{tabular}[c]{@{}l@{}}I'm attending a meeting in Japan and would like to give a gift to my hosts.\\ What should I consider to ensure my gesture is well-received?\end{tabular} \\ \hline
\textbf{Actor Behavior} & Behavior of the actor & Offer a gift wrapped in traditional Japanese style as a gesture of appreciation \\ \hline
\textbf{Recipient} & Recipient of the action & Japanese business hosts \\ \hline
\textbf{Relation} & Relation between the actor and the recipient & Business partners \\ \hline
\textbf{Recipient Behavior} & Behavior of the recipient & Receive the gift with both hands and show appreciation

\end{tabular}
}
\captionsetup{font=footnotesize}
\caption{\footnotesize An example of a social commonsense question from CultureBank.}
\label{fig:culbank_example}
\end{table}

Each seed instance follows the same 4-step process as described in previous section to generate the final \textit{context-implicit commonsense question}. However, minor differences are introduced to adapt to CultureBank data, including:

\textbf{Seed Data Filtering:}  In addition to the 3 criteria used for \textbf{mSCoRe-G}, we introduce an additional criterion - \textbf{multiculture-ness} - for filtering social situations (detailed description provided in Appendix \ref{app:judge}). The aim is to select situations that involve the most culturally distinctive elements, allowing us to evaluate models' understanding of diverse cultural contexts and associated commonsense knowledge.

\textbf{Structured Reasoning Generation:} Before generating the context and reasoning process, LLM needs to generate the QA pair first.
This acts as the seed QA pair from mCSQA in the previous section and goes through the same procedure.

\vspace{-1mm}
\subsubsection{Dataset Statistic}
\vspace{-1mm}
\textbf{mSCoRe-G} covers 5 languages including English, German, French, Japanese, and Chinese. For each language, we create 200 examples ranging from level 0 (original QA pair) to level 3 (3 steps of expansion). This results in 800 examples per language.
For \textbf{mSCoRe-S}, we similarly create 200 examples for each source (TikTok and Reddit).
In total, \textbf{mSCoRe} contains 5,600 instances (4000 for general commonsense and 1600 for social commonsense). 
Detailed examples at different complexity levels are provided in Appendix \ref{app:prompt}.


\vspace{-1mm}
\section{Experiments}
\vspace{-2mm}

\subsection{Experiment Setup}

We conduct comprehensive evaluations using a diverse set of state-of-the-art multilingual language models, selected to represent different approaches to model development and training. Our evaluation considers three key dimensions: model availability, parameter scale, and training methodology.
The models evaluated in our study include:

\textbf{GPT-4o} \citep{gpt4o}: A general-purpose LLM representing the current state-of-the-art LLM, trained on large-scale multimodal data from diverse sources.

\begin{table*}[htbp]
\centering
\addtolength{\abovecaptionskip}{-2mm}
\addtolength{\belowcaptionskip}{-2mm}

\resizebox{0.99\textwidth}{!}{

\begin{tabular}{l|cccc|cccc|cccc|cccc|cccc|cccc}
\multicolumn{1}{c|}{} & \multicolumn{4}{c|}{English} & \multicolumn{4}{c|}{German} & \multicolumn{4}{c|}{French} & \multicolumn{4}{c|}{Chinese} & \multicolumn{4}{c|}{Japanese} & \multicolumn{4}{c}{Average} \\ \cline{2-25} 
\multicolumn{1}{c|}{\multirow{-2}{*}{\begin{tabular}[c]{@{}c@{}}General\\ Commonsense\end{tabular}}} & \cellcolor[HTML]{D9D9D9}L0 & \cellcolor[HTML]{D9D9D9}L1 & \cellcolor[HTML]{D9D9D9}L2 & \cellcolor[HTML]{D9D9D9}L3 & \cellcolor[HTML]{D9D9D9}L0 & \cellcolor[HTML]{D9D9D9}L1 & \cellcolor[HTML]{D9D9D9}L2 & \cellcolor[HTML]{D9D9D9}L3 & \cellcolor[HTML]{D9D9D9}L0 & \cellcolor[HTML]{D9D9D9}L1 & \cellcolor[HTML]{D9D9D9}L2 & \cellcolor[HTML]{D9D9D9}L3 & \cellcolor[HTML]{D9D9D9}L0 & \cellcolor[HTML]{D9D9D9}L1 & \cellcolor[HTML]{D9D9D9}L2 & \cellcolor[HTML]{D9D9D9}L3 & \cellcolor[HTML]{D9D9D9}L0 & \cellcolor[HTML]{D9D9D9}L1 & \cellcolor[HTML]{D9D9D9}L2 & \cellcolor[HTML]{D9D9D9}L3 & \cellcolor[HTML]{D9D9D9}L0 & \cellcolor[HTML]{D9D9D9}L1 & \cellcolor[HTML]{D9D9D9}L2 & \cellcolor[HTML]{D9D9D9}L3 \\ \hline
GPT-4o & \textbf{80.5} & 70.0 & \textbf{72.5} & \textbf{71.5} & \textbf{75.0} & \textbf{68.5} & {\ul \textbf{71.0}} & 67.5 & 78.0 & {\ul \textbf{74.0}} & \textbf{70.0} & 63.5 & \cellcolor[HTML]{FFFFFF}{\ul \textbf{80.5}} & \cellcolor[HTML]{FFFFFF}{\ul \textbf{78.5}} & \cellcolor[HTML]{FFFFFF}{\ul \textbf{72.5}} & \textbf{65.5} & \textbf{82.0} & \textbf{83.5} & \cellcolor[HTML]{FFFFFF}{\ul \textbf{79.5}} & {\ul \textbf{79.5}} & {\ul \textbf{79.2}} & \textbf{74.9} & {\ul \textbf{73.1}} & \textbf{69.5} \\
o1 & {\ul \textbf{82.5}} & \textbf{73.5} & \cellcolor[HTML]{FFFFFF}{\ul \textbf{75.0}} & {\ul \textbf{72.0}} & \textbf{75.0} & 67.5 & 63.0 & 67.5 & \cellcolor[HTML]{FFFFFF}{\ul \textbf{80.5}} & \textbf{72.5} & {\ul \textbf{71.5}} & 61.0 & 64.5 & 63.0 & 56.0 & 53.0 & 80.5 & 80.0 & 77.0 & 73.0 & 76.6 & 71.3 & 68.5 & 65.3 \\
o1-mini & 76.5 & 70.5 & 65.5 & 63.5 & 69.5 & 66.0 & \textbf{69.5} & 64.5 & 71.5 & 64.5 & 59.5 & 55.0 & 71.0 & 63.0 & 60.0 & 51.5 & 77.5 & 75.0 & 68.5 & 66.5 & 73.2 & 67.8 & 64.6 & 60.2 \\ \hline
LLaMA-3.3-70B & 78.5 & {\ul \textbf{75.0}} & 69.0 & 70.0 & {\ul \textbf{75.5}} & {\ul \textbf{72.5}} & 68.0 & {\ul \textbf{73.0}} & \textbf{78.5} & 72.0 & 67.0 & \textbf{64.0} & \textbf{80.0} & \textbf{74.5} & \textbf{70.5} & {\ul \textbf{67.0}} & \textbf{82.0} & \cellcolor[HTML]{FFFFFF}{\ul \textbf{85.5}} & \textbf{76.5} & \textbf{78.0} & \textbf{78.9} & {\ul \textbf{75.9}} & \textbf{70.2} & {\ul \textbf{70.4}} \\
LLaMA-3.1-8B & 23.0 & 22.5 & 21.5 & 21.5 & 73.0 & 65.5 & 63.0 & 61.0 & 69.5 & 61.0 & 54.5 & 52.0 & 60.0 & 52.0 & 46.0 & 43.0 & 17.5 & 18.5 & 17.0 & 17.5 & 48.6 & 43.9 & 40.4 & 39.0 \\ \hline
R1-70B & 79.5 & 70.5 & 69.5 & 69.0 & 73.0 & 67.0 & 67.0 & \textbf{70.0} & 76.0 & 71.5 & 69.5 & {\ul \textbf{64.5}} & 75.0 & 70.5 & 61.0 & 65.0 & \cellcolor[HTML]{FFFFFF}{\ul \textbf{83.0}} & 79.5 & 72.0 & 73.5 & 77.3 & 71.8 & 67.8 & 68.4 \\
R1-8B & 67.5 & 62.0 & 62.0 & 55.0 & 67.5 & 58.0 & 61.0 & 55.5 & 58.0 & 45.0 & 44.0 & 43.5 & 69.0 & 62.0 & 51.5 & 58.5 & 61.5 & 57.0 & 59.0 & 53.5 & 64.7 & 56.8 & 55.5 & 53.2 \\ \hline
Aya-32B & 77.5 & 67.0 & 66.5 & 66.0 & 70.5 & 65.5 & 66.5 & 66.0 & 76.5 & 69.0 & 65.0 & 60.5 & 78.0 & 67.0 & 64.0 & 60.0 & 79.5 & 80.5 & 70.0 & 72.5 & 76.4 & 69.8 & 66.4 & 65.0
\end{tabular}

}
\captionsetup{font=footnotesize}
\caption{\footnotesize Accuracy comparison of models on \textbf{mSCoRe-G} from complexity level 0 (L0) to 3 (L3).}
\label{tab:general_main}
\end{table*}

OpenAI \textbf{o1} \citep{o1:24}: A reasoning-reinforced model based on GPT-4o, specifically optimized for complex problem-solving tasks through an additional training phase utilizing data curated for chain-of-thought reasoning.

\textbf{LLaMA-3.3-70B} and \textbf{LLaMA-3.1-8B} \citep{llama3}: Two open-sourced LLMs representing different parameter scales, trained on publicly available sources spanning various domains, allowing us to analyze the impact of model size on reasoning capabilities.

Distilled DeepSeek-R1 (\textbf{R1-70B} and \textbf{R1-8B}) \citep{r1:25}: A reasoning-focused model derived from the LLaMA architecture, distilled using samples generated by the large-scale LRM DeepSeek-R1.

\textbf{Aya-32B} \citep{aya32}: A universal multilingual model trained on data from 200 languages, providing insights into broad multilingual LLM reasoning capabilities.

For evaluation, we employ a consistent prompt for all models, providing the proposed reasoning skill taxonomy (section \ref{3.1-taxonomy}) with step-by-step instructions to generate the desired reasoning process before answering. Further experimental details are in Appendix \ref{app:exp}. 

\begin{wraptable}{R}{0.56\textwidth}
\centering
    \vspace{-5mm}
\addtolength{\abovecaptionskip}{-2mm}
\addtolength{\belowcaptionskip}{-3mm}
\resizebox{0.56\textwidth}{!}{

\begin{tabular}{l|cccc|cccc|cccc}
 & \multicolumn{4}{c|}{TikTok} & \multicolumn{4}{c|}{Reddit} & \multicolumn{4}{c}{Average} \\ \cline{2-13} 
\multirow{-2}{*}{\begin{tabular}[c]{@{}c@{}}Social\\ Commonsense\end{tabular}} & \cellcolor[HTML]{D9D9D9}L0 & \cellcolor[HTML]{D9D9D9}L1 & \cellcolor[HTML]{D9D9D9}L2 & \cellcolor[HTML]{D9D9D9}L3 & \cellcolor[HTML]{D9D9D9}L0 & \cellcolor[HTML]{D9D9D9}L1 & \cellcolor[HTML]{D9D9D9}L2 & \cellcolor[HTML]{D9D9D9}L3 & \cellcolor[HTML]{D9D9D9}L0 & \cellcolor[HTML]{D9D9D9}L1 & \cellcolor[HTML]{D9D9D9}L2 & \cellcolor[HTML]{D9D9D9}L3 \\ \hline
GPT-4o & \textbf{71.0} & \textbf{69.0} & 62.5 & \textbf{63.5} & 75.0 & 67.0 & \textbf{68.5} & \textbf{69.5} & 73.0 & 68.0 & \textbf{65.5} & \textbf{66.5} \\
o1 & 69.5 & \textbf{69.0} & \textbf{63.5} & 61.5 & \textbf{77.0} & \textbf{71.0} & 67.5 & 69.0 & \textbf{73.3} & \textbf{70.0} & \textbf{65.5} & 65.3 \\
o1-mini & 63.5 & 62.5 & 53.0 & 59.5 & 72.5 & 62.0 & 62.0 & 59.0 & 68.0 & 62.3 & 57.5 & 59.3 \\ \hline
LLaMA-3.3-70B & {\ul \textbf{80.0}} & {\ul \textbf{75.0}} & {\ul \textbf{73.5}} & {\ul \textbf{73.0}} & {\ul \textbf{83.5}} & {\ul \textbf{76.5}} & {\ul \textbf{80.0}} & {\ul \textbf{76.5}} & {\ul \textbf{81.8}} & {\ul \textbf{75.8}} & {\ul \textbf{76.8}} & {\ul \textbf{74.8}} \\
LLaMA-3.1-8B & 30.5 & 29.5 & 29.5 & 29.5 & 27.0 & 27.0 & 27.0 & 27.0 & 28.8 & 28.3 & 28.3 & 28.3 \\ \hline
R1-70B & 68.5 & 65.5 & 62.0 & 62.5 & 73.5 & 67.0 & 67.5 & 67.5 & 71.0 & 66.3 & 64.8 & 65.0 \\
R1-8B & 64.0 & 60.0 & 56.5 & 52.5 & 65.0 & 58.5 & 60.5 & 61.0 & 64.5 & 59.3 & 58.5 & 56.8 \\ \hline
Aya-32B & 68.0 & 64.0 & 61.0 & 60.5 & 71.0 & 57.5 & 63.0 & 59.5 & 69.5 & 60.8 & 62.0 & 60.0
\end{tabular}

}
\captionsetup{font=footnotesize}
\caption{ \footnotesize Accuracy comparison of models on \textbf{mSCoRe-S}.}
\label{tab:social_main}
\end{wraptable}


\vspace{-1mm}
\subsection{Main Results}
\vspace{-1mm}
Tables \ref{tab:general_main} and \ref{tab:social_main} present our main results on \textbf{mSCoRe-G} and \textbf{mSCoRe-S}, respectively. 
Overall, we observe a consistent pattern across all models where performance declines as complexity level increases.
For \textbf{mSCoRe-G}, GPT-4o achieves the highest overall accuracy on general commonsense reasoning across all languages and complexity levels.
While this can be an artifact of the benchmark creation process where GPT-4o was used for data generation, LLaMA-3.3-70B results are very close to GPT-4o. 
Furthermore, the open-source model significantly outperforms others on social commonsense reasoning (over 5\% average improvement across all levels and domains).


\textbf{Multilingual and Cultural Results: } Performance is generally similar across languages in \textbf{mSCoRe-G}. This may be due to all languages in the seed dataset mCSQA being medium to high resource languages. Future work should explore other seed datasets with more low-resource languages.
For social commonsense reasoning in \textbf{mSCoRe-S}, most models perform better on Reddit-sourced questions than TikTok-sourced ones. This could be attributed to Reddit containing more content on general ``Community and Cultural Exchange'', whereas TikTok focuses more on daily life ``personal" aspects like ``Social Norms and Etiquette". This suggests that LLMs might still struggle with more personalized problems, as noted in \cite{survey_ernest:23}.
Unexpectedly, despite being trained on 200 diverse languages, the most multilingual model Aya-32B does not perform very well, even in cultural social commonsense benchmark.

\textbf{Model Scale: } 
We compare models with different parameter counts, from the 8B and 70B parameters open-source models (LLaMA and R1), to the colossal-scale (hundreds of billions of parameters) closed-source LLMs (GPT-4o and o1). 
Larger models generally perform better across both benchmarks. 
The performance gap between 70B and 8B versions was substantial in most cases. 
However, we observe diminishing returns when moving from 70B to colossal-scale LLMs. 
This finding suggests it takes more than simple parameter scaling to solve commonsense reasoning, especially in understanding social interactions and cultural norms.

\begin{figure*}[ht!]
\addtolength{\belowcaptionskip}{-3mm}
\addtolength{\belowcaptionskip}{-4mm}
\begin{center} 
\includegraphics[width=0.99\textwidth]{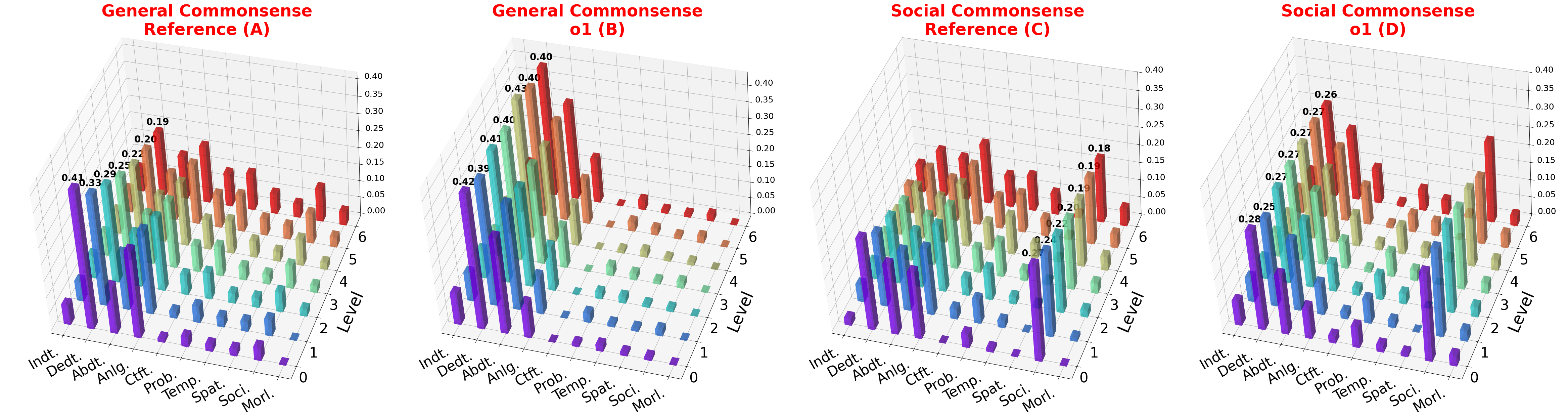}
\captionsetup{font=footnotesize}
\caption{\footnotesize The distribution of reasoning skills of reference reasoning process (\textbf{A} and \textbf{C}), and o1's reasoning process (\textbf{B} and \textbf{D}), as question complexity increases from complexity level 0 to 6.}

\label{fig:3d_skill}
\end{center}
\end{figure*}

\textbf{Reasoning-reinforced Training: }
 We compare general instruction-tuned models (GPT-4o, LLaMA) with reasoning-reinforced fine-tuned models (o1 and R1).
 While the state-of-the-art LRM o1 performs best in English, it lags behind other general LLMs like GPT-4o and LLaMA-3.3-70B in other languages.
 This suggests that reasoning-reinforced training might decrease commonsense reasoning ability, likely due to the highly specialized training data for more complex tasks like coding and math. Interestingly, LLaMA-3.1-8B fails the task in English and Japanese, but R1-8B performs normally, indicating that reasoning-reinforced training helps smaller-scale models understand tasks better.

\vspace{-1mm}
\section{Analysis}
\vspace{-2mm}



\begin{wraptable}{R}{0.6\textwidth}
\centering
    \vspace{-13mm}
\addtolength{\abovecaptionskip}{-3mm}
\addtolength{\belowcaptionskip}{-2mm}
\resizebox{0.6\textwidth}{!}{

\begin{tabular}{l|ccccccc|ccccccc}
\multicolumn{1}{l|}{} & \multicolumn{7}{c|}{General (English)} & \multicolumn{7}{c}{Social (Average)} \\ \cline{2-15} 
\multicolumn{1}{l|}{\multirow{-2}{*}{}} & \cellcolor[HTML]{D9D9D9}L0 & \cellcolor[HTML]{D9D9D9}L1 & \cellcolor[HTML]{D9D9D9}L2 & \cellcolor[HTML]{D9D9D9}L3 & \cellcolor[HTML]{D9D9D9}L4 & \cellcolor[HTML]{D9D9D9}L5 & \cellcolor[HTML]{D9D9D9}L6 & \cellcolor[HTML]{D9D9D9}L0 & \cellcolor[HTML]{D9D9D9}L1 & \cellcolor[HTML]{D9D9D9}L2 & \cellcolor[HTML]{D9D9D9}L3 & \cellcolor[HTML]{D9D9D9}L4 & \cellcolor[HTML]{D9D9D9}L5 & \cellcolor[HTML]{D9D9D9}L6 \\ \cline{1-15} 
GPT-4o & \textbf{80.5} & 70.0 & \textbf{72.5} & \textbf{71.5} & {\ul \textbf{72.0}} & {\ul \textbf{70.0}} & \textbf{68.0} & 73.0 & 68.0 & \textbf{65.5} & \textbf{66.5} & \textbf{66.3} & \textbf{67.3} & \textbf{66.0} \\
o1 & {\ul \textbf{82.5}} & \textbf{73.5} & {\ul \textbf{75.0}} & {\ul \textbf{72.0}} & \textbf{70.0} & \textbf{69.5} & 67.0 & \textbf{73.3} & \textbf{70.0} & \textbf{65.5} & 65.3 & 62.8 & 60.8 & 63.0 \\ \hline
LLaMA-3.3-70B & 78.5 & {\ul \textbf{75.0}} & 69.0 & 70.0 & 68.5 & 68.5 & {\ul \textbf{69.0}} & {\ul \textbf{81.8}} & {\ul \textbf{75.8}} & {\ul \textbf{76.8}} & {\ul \textbf{74.8}} & {\ul \textbf{75.5}} & {\ul \textbf{72.5}} & {\ul \textbf{73.3}} \\
LLaMA-3.1-8B & 23.0 & 22.5 & 21.5 & 21.5 & 20.5 & 20.5 & 20.5 & 28.8 & 28.3 & 28.3 & 28.3 & 28.0 & 28.0 & 28.0 \\ \hline
Deepseek-70B & 79.5 & 70.5 & 69.5 & 69.0 & 69.0 & 69.0 & 65.5 & 71.0 & 66.3 & 64.8 & 65.0 & 60.3 & 62.8 & 65.8 \\
Deepseek-8B & 67.5 & 62.0 & 62.0 & 55.0 & 57.0 & 59.5 & 55.0 & 64.5 & 59.3 & 58.5 & 56.8 & 59.3 & 59.8 & 61.5 \\ \hline
Aya-32B & 77.5 & 67.0 & 66.5 & 66.0 & 58.0 & 60.0 & 54.5 & 69.5 & 60.8 & 62.0 & 60.0 & 57.8 & 56.5 & 54.0
\end{tabular}
}
\captionsetup{font=scriptsize}
\caption{\scriptsize Performance comparison from complexity level 0 to level 6.}
\label{tab:scale_table}

\end{wraptable}

\subsection{Complexity Scaling Results}  
\vspace{-2mm}
To further understand model capacity against scaling question complexity, we expand our results for \textbf{mSCoRe-G} (English) and \textbf{mSCoRe-S} to complexity level 6.
As shown in Table. \ref{tab:scale_table}, every model accuracy continues to decline to L6.
The most significant performance drop occurs between L0 and L2, indicating that even relatively simple complexity scaling introduces substantial challenges for LLMs.
At higher difficulty levels (L3 to L6), the rate of degradation slows down considerably. 
This plateau suggests that our current approach to scaling complexity through additional context and reasoning steps may reach a saturation point. This could indicate that the multiple-choice question-answer format itself imposes certain limitations on how effectively task difficulty can be scaled. Alternative task formulations that require more sophisticated forms of reasoning beyond the current design might be necessary to create more discriminative benchmarks for future, more capable models.

\vspace{-1mm}
\subsection{Skill Type Utilization}
\vspace{-1mm}
To better understand how models employ different reasoning skills across varying complexity levels, Fig. \ref{fig:3d_skill} visualizes the distribution of reasoning skills used in both the reference reasoning processes (from our benchmark creation) and the output reasoning processes generated by o1.

For \textit{general} commonsense, both reference and model-generated reasoning primarily utilize logical reasoning skills, with deductive reasoning being most common. However, the reference distribution shows greater diversification of skills at higher complexity levels, incorporating more contextual reasoning (especially analogical and probabilistic reasoning). In contrast, models like o1 remains heavily dependent on deductive reasoning across all complexity levels.
For \textit{social} commonsense, the reference distribution shows more balanced utilization of skills from all three categories, with social and ethical reasoning becomes progressively more important for higher-level questions. While o1 model incorporates some social reasoning skills, it still over-relies on logical reasoning for scenarios where social and contextual reasoning would be more appropriate.
Overall, results reveal significant limitations in o1's ability to adapt its reasoning strategy. The rigid reasoning pattern likely explains the model's performance decrease on higher complexity questions, highlighting the need for more balanced reasoning-reinforced training approaches.


\begin{wraptable}{R}{0.45\textwidth}

\centering
    \vspace{-10mm}
\addtolength{\abovecaptionskip}{-2mm}
\addtolength{\belowcaptionskip}{-5mm}
\resizebox{0.45\textwidth}{!}{

\begin{tabular}{l|cccc|cccc}
\multicolumn{1}{l|}{} & \multicolumn{4}{c|}{General} & \multicolumn{4}{c}{Social} \\ \cline{2-9} 
\multicolumn{1}{l|}{\multirow{-2}{*}{}} & \cellcolor[HTML]{D9D9D9}L0 & \cellcolor[HTML]{D9D9D9}L1 & \cellcolor[HTML]{D9D9D9}L2 & \cellcolor[HTML]{D9D9D9}L3 & \cellcolor[HTML]{D9D9D9}L0 & \cellcolor[HTML]{D9D9D9}L1 & \cellcolor[HTML]{D9D9D9}L2 & \cellcolor[HTML]{D9D9D9}L3 \\ \hline
o1 & 76.6 & {\ul \textbf{71.3}} & {\ul \textbf{68.5}} & 65.3 & {\ul \textbf{73.3}} & {\ul \textbf{70.0}} & {\ul \textbf{65.5}} & {\ul \textbf{65.3}} \\
o1-mini & 73.2 & 67.8 & 64.6 & 60.2 & 68.0 & 62.3 & 57.5 & \textbf{59.3} \\ \hline
cot-o1 & 75.9 & 69.3 & 66.2 & 61.3 & 63.3 & 49.3 & 44.5 & 40.3 \\
cot-o1-mini & 71.7 & 65.2 & 60.2 & 57.8 & 60.8 & 51.5 & 46.5 & 45.3 \\ \hline
logical-o1 & \textbf{77.3} & 72.1 & 66.3 & \textbf{65.6} & \textbf{72.8} & \textbf{64.3} & \textbf{59.8} & 58.3 \\
logical-o1-mini & 73.9 & 68.3 & 62.5 & 62.2 & 64.8 & 59.8 & 59.3 & 50.5 \\ \hline
general-o1 & {\ul \textbf{77.7}} & \textbf{69.9} & \textbf{67.5} & {\ul \textbf{65.8}} & 69.3 & 54.5 & 51.5 & 48.3 \\
general-o1-mini & 73.3 & 67.7 & 61.4 & 59.6 & 66.8 & 61.8 & 57.0 & 52.8
\end{tabular}
}
\captionsetup{font=scriptsize}
\caption{\scriptsize Results for Different Reasoning Skill Taxonomies.}
\label{tab:prompt_skill}
\end{wraptable}

\vspace{-1mm}
\subsection{Different Reasoning Skill Taxonomies}
\vspace{-1mm}
We investigate how models adapt to different reasoning taxonomies, including: 
(1) \textbf{Chain-of-Thought (CoT)} - Standard chain-of-thought, not requiring skill identification, 
(2) \textbf{Logical} - Only using logical reasoning skills (deductive, inductive and abductive)
(3) \textbf{General} - Each reasoning step is categorized into one of the three general categories (logical, contextual, and social).

Table \ref{tab:prompt_skill} shows the average accuracies of o1 and o1-mini for each setting.
Interestingly, despite requiring models to distinguish between more skill types,
our proposed fine-grained taxonomy yields the best results.
As expected from our previous analysis, the \textbf{Logical}-only approach performs relatively well on general commonsense tasks but worse on social tasks.
The \textbf{General} setting also under-performs ours, suggesting that granularity of skill identification benefits commonsense reasoning by encouraging models to consider a broader range of reasoning approaches rather than defaulting to familiar patterns.
Finally, \textbf{CoT} performs notably worse than all structured skill-based approaches, especially for social commonsense at higher complexity levels. This demonstrates that reasoning without explicit skill categorization may be insufficient for more complex commonsense situations.

\begin{wrapfigure}{R}{0.5\textwidth}
\vspace{-10mm}
\addtolength{\abovecaptionskip}{-3mm}
\addtolength{\belowcaptionskip}{-5mm}
\begin{center} 
\includegraphics[width=0.48\textwidth]{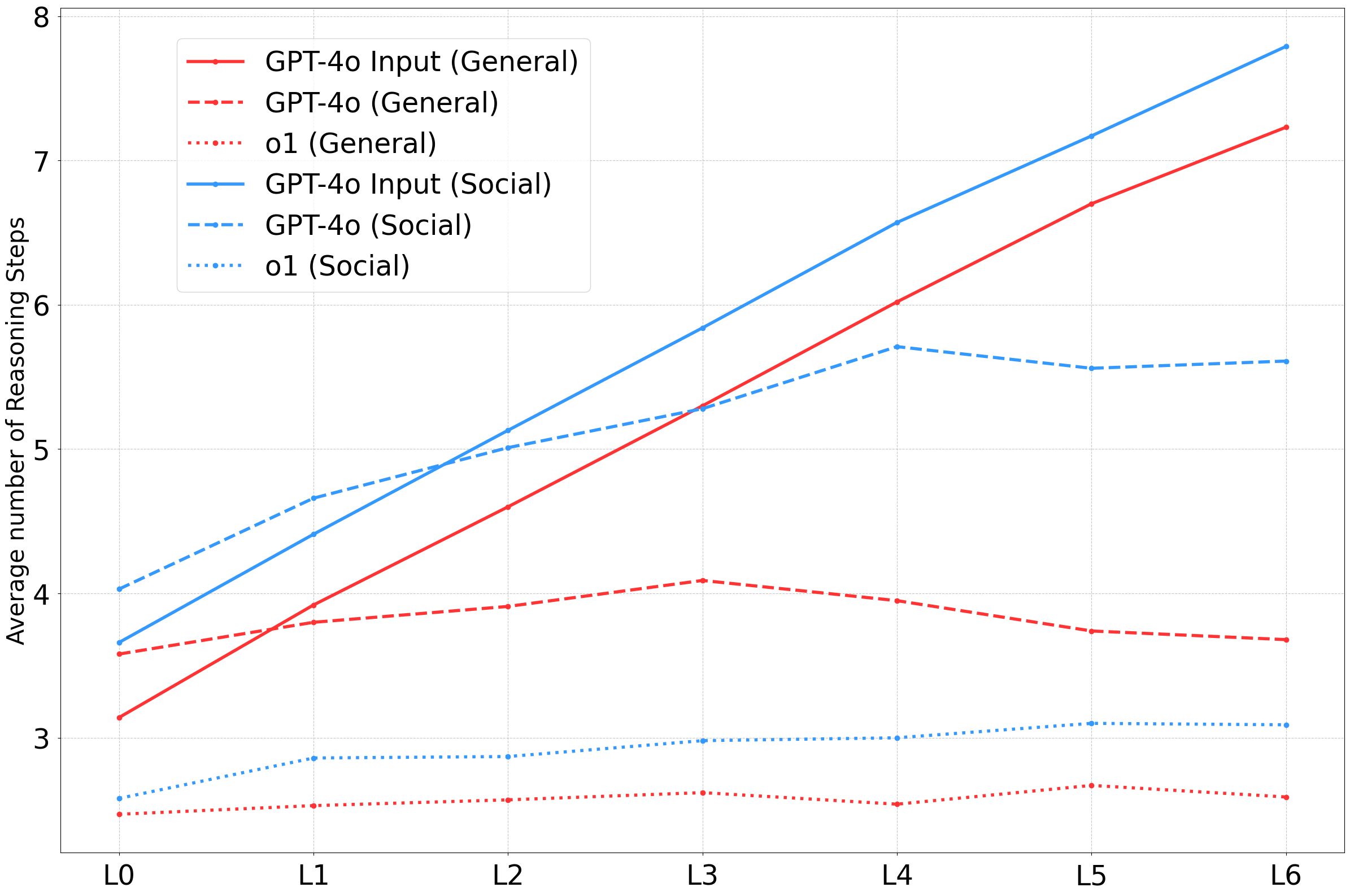}
\captionsetup{font=scriptsize}
\caption{\scriptsize Average number of reasoning steps in the reasoning processes of \textbf{mSCoRe}(straight), GPT-4o (barred), and o1 (dotted). 
}
\label{fig:skill_effi}
\end{center}
\end{wrapfigure}

\vspace{-1mm}
\subsection{Reasoning Efficiency}
\vspace{-1mm}
To examine the relationship between reasoning efficiency and task complexity across different models, Fig. \ref{fig:skill_effi} visualizes the average number of steps of the reasoning processes of \textbf{mSCoRe} and GPT-4o and o1's answers across different complexity levels.

The reference reasoning processes show a clear linear increase in reasoning steps as complexity level increases, with social commonsense reasoning requiring more steps than general commonsense at each level.
GPT-4o's reasoning processes show a similar upward trend but with a more gradual slope, whereas o1's reasoning processes maintain a nearly constant step count (around 3 steps) regardless of task complexity.  
The results indicate higher-level complexities require more steps, and current model are unable to reason longer, unless explicitly forced to (such as in the Complexity Expansion step described in Section \ref{3.2-mscore}).
This is similar to the number of reasoning tokens (more steps is equivalent to more tokens) used in test-time scaling research recently introduced in \cite{s1:25}.
These findings indicate that adapting reasoning depth dynamically based on task demands is likely crucial for sustaining performance as complexity escalates.






\vspace{-3mm}
\section{Conclusion}
\vspace{-3mm}
We introduce \textbf{mSCoRe} - Multilingual and Scalable Benchmark for Skill-based Commonsense Reasoning, a novel evaluation framework designed to address critical gaps in existing benchmarks for commonsense reasoning. By integrating multilingual and diverse cultural coverage, a fine-grained reasoning skill taxonomy, and a dynamic complexity scaling mechanism, \textbf{mSCoRe} provides a comprehensive platform for systematically evaluating not only the accuracy but also skill utilization and efficiency of LLMs' commonsense reasoning process.
Extensive experiments on eight state-of-the-art LLMs reveal that current models still consistently struggled with higher complexity levels and culturally nuanced social commonsense scenarios.
Our analysis highlights several promising directions for improvement, including more robust training methodologies to enhance model's reasoning skill utilization and efficiency.
Additionally, \textbf{mSCoRe} provides a framework for subsequent benchmarks to scale with the rapid development of LLMs in the future.

\bibliography{colm2025_conference}
\bibliographystyle{colm2025_conference}

\clearpage
\appendix
\section{Experimental Details} \label{app:exp}

\paragraph{\textbf{LLMs Details: }}
For closed commercial LLMs (GPT-4o, o1 and o1-mini), we query responses from the models using OpenAI Chat Completions API\footnote{https://platform.openai.com/docs/guides/text-generation}, with temperatures set to 0 for deterministic outputs.
Open source models (
Deepseek R1-70B\footnote{https://huggingface.co/deepseek-ai/DeepSeek-R1-Distill-Llama-70B} and 
R1-8B\footnote{https://huggingface.co/deepseek-ai/DeepSeek-R1-Distill-Llama-8B}, 
LLaMA-3.3-70B\footnote{https://huggingface.co/meta-llama/Llama-3.3-70B-Instruct} and 
LLaMA-3.1-8B\footnote{https://huggingface.co/meta-llama/Llama-3.1-8B}, 
Aya-32B\footnote{https://huggingface.co/CohereForAI/aya-expanse-32b}
) are run using 2 NVIDIA A100 80GB GPUs.
PyTorch 2.1.2\footnote{https://pytorch.org/get-started/pytorch-2.0/} and Huggingface-Transformer 4.42.3 \footnote{https://github.com/huggingface/transformers} are used to implement the models. \\


\paragraph{\textbf{Source code with specification of all dependencies, including external libraries: }} Our data and source code will be released upon acceptance of the paper.

\subsection{Reasoning Skill Details} \label{app:skill}
We provide detail descriptions with abstract and concrete example for each of our reasoning skills in Fig. \ref{fig:app_reason_skill}.
Abstract examples are generalized representation that uses variables or placeholders to illustrate a pattern or principle of the reasoning skills.
In contrast, concrete examples are the application of the corresponding reasoning skills to a specific real-world scenario.

\section{Data Generation Details}

\subsection{LLM-Judge} \label{app:judge}

\paragraph{Judge Model} 
We utilize Flow Judge, a general LLM-as-a-judge model developed by Flow AI\footnote{https://www.flow-ai.com/blog/flow-judge}. Flow Judge is an open-source 3.8B parameter language model designed for LM-based evaluations, offering high performance and accuracy comparable to much larger models like GPT-4o and Claude 3.5 Sonnet. It is trained on evaluation data across various domains to supports custom evaluation criteria, multiple scoring scales, qualitative feedback, and produces structured evaluation outputs.

\paragraph{Scoring Metrics for Seed Data Filtering step:} 
We provide the full rubrics used for Seed Data Filtering step for \textbf{mSCoRe-G} and \textbf{mSCoRe-S} in Fig. \ref{fig:app_rubric_mcsqa} and \ref{fig:app_rubric_culbank}, correspondingly.

\section{Prompt Details} \label{app:prompt}

We provide here all of the prompt templates in full version used in our experiments.
Fig. \ref{fig:app_prompt_reason_gen_general}, \ref{fig:app_prompt_complex_exp}, and \ref{fig:app_prompt_com_impl} presents the prompt of Structured Reasoning Generation, Data Complexity Scaling and Commonsense Implicitation steps in our data generation process for \textbf{mSCoRe-G}.
Additionally, Fig. \ref{fig:app_prompt_reason_gen_social} presents the prompt  of Structured Reasoning Generation step for \textbf{mSCoRe-S} (The other two steps remain the same between the two subsets).

We provide an example of complexity level 0 to 3 for \textbf{mSCoRe-G} and \textbf{mSCoRe-S} in Fig. \ref{fig:app_general_example} and \ref{fig:app_social_example}, correspondingly.

\begin{figure*}[]
\captionsetup{skip=2pt}
\centering \tiny \obeylines
\begin{tcolorbox}[width=0.97\textwidth]
\begin{Verbatim}[breaklines, breaksymbolleft=]
{
"inductive_reasoning": {
    "short_description": "Drawing general conclusions from specific observations.",
    "long_description": "Inductive reasoning is a method of drawing general conclusions from specific observations. Unlike deductive reasoning, which starts with general premises to reach specific conclusions, inductive reasoning begins with detailed facts and builds up to broader generalizations or theories. This approach is commonly used in scientific research, where repeated experiments and observations lead to the formulation of overarching principles or hypotheses ",
    "abstract_example": "After witnessing several instances where Event A_1 leads to Event A_2, you infer that Event A_n will similarly result in Event A_2 in future occurrences",
    "concrete_example": "After witnessing several instances where the weather forecast predicts rain, you infer that rain will likely continue to fall in the future"
},
"deductive_reasoning": {
    "short_description": "Deriving specific conclusions from general premises.",
    "long_description": "Deductive reasoning involves deriving specific conclusions from general premises. It ensures that if the premises are true and the reasoning is valid, the conclusion must also be true. Deductive logic is fundamental in fields that require rigorous proof, such as mathematics and formal sciences.",
    "abstract_example": "Given the premise that All X are Y, and knowing that Object x_1 is an X, you deduce that Object x_1 must also be a Y.",
    "concrete_example": "Given All birds have feathers. A sparrow is a bird. Therefore, a sparrow has feathers"
},
"abductive_reasoning": {
    "short_description": "Forming hypotheses to explain observations.",
    "long_description": "Abductive reasoning is the process of forming hypotheses to explain observations. It starts with an incomplete set of observations and proceeds to the likeliest possible explanation. Unlike deductive and inductive reasoning, abductive reasoning seeks the simplest and most plausible explanation for a given set of facts, often leading to the generation of new theories or hypotheses.",
    "abstract_example": "Observing Event B, you hypothesize that Reason 2 is the most plausible explanation among several possible causes.",
    "concrete_example": "You wake up and see that the street is wet. The most likely explanation is that it rained last night."
},
"analogical_reasoning": {
    "short_description": "Drawing parallels between similar situations to infer conclusions.",
    "long_description": "Analogical reasoning involves drawing parallels between similar situations to infer conclusions. By comparing two objects or systems that share certain characteristics, one can infer that they may share additional, unobserved properties. This form of reasoning is widely used in problem-solving, scientific discovery, and legal reasoning to transfer knowledge from a known domain (source) to an unknown domain (target). Analogical reasoning is also used in everyday life to make inferences about the similarities between objects or situations.",
    "abstract_example": "Think of Situation C_a, where Component C_a_1 interacts with Component C_a_2 in a specific way. You encounter Situation C_b with Component C_b_1 and Component C_b_2, and infer that Component C_b_1 and Component C_b_2 will interact similarly in Situation C_a.",
    "concrete_example": "Just as a gardener waters plants to help them grow, a teacher provides knowledge and guidance to help students develop."
},
"counterfactual_reasoning": {
    "short_description": "Considering alternative scenarios and outcomes that did not happen.",
    "long_description": "Counterfactual reasoning entails considering alternative scenarios and outcomes that did not occur. It involves imagining 'what might have happened' under different circumstances, which is useful for understanding causality, evaluating decisions, and planning future actions. Counterfactual reasoning is often used in fields such as philosophy, psychology, and business to explore the potential consequences of different choices or actions.",
    "abstract_example": "Reflecting on Condition X that did not occur, you imagine that if it had, Outcome Y might have replaced Outcome Z.",
    "concerte_example": "If you had left the house five minutes earlier, you would have caught the bus on time."
},
"probabilistic_reasoning": {
    "short_description": "Applying principles of probability to make inferences under uncertainty.",
    "long_description": "Probabilistic reasoning involves applying principles of probability to make inferences under uncertainty. It enables individuals to assess the likelihood of different outcomes and make informed decisions based on the probability of various events occurring. This type of reasoning is crucial in fields like statistics, risk assessment, and artificial intelligence.",
    "abstract_example": "Evaluating that Option A has a higher probability (P(A) > P(B)) of success than Option B, you decide to choose Option A.",
    "concrete_example": "There is a 70% chance of rain tomorrow, so you decide to carry an umbrella when you go out."
},
"temporal_reasoning": {
    "short_description": "Understanding sequences and durations of events.",
    "long_description": "Temporal reasoning is the ability to understand and reason about the sequence and duration of events over time. It involves comprehending time-specific data, such as the order of events, how long events last, and the relationships between different time points. Temporal reasoning is essential in areas like scheduling, planning, and understanding narratives.",
    "abstract_example": "Planning your day, you schedule Event T_1 to occur before Event T_2, ensuring the correct sequence of activities.",
    "concrete_example": "You observe that the sun will rise in the morning and set in the evening. You infer that the moon will rise and set at the same time."
},
\end{Verbatim}
\end{tcolorbox}
\end{figure*}

\begin{figure*}[]
\captionsetup{skip=2pt}
\centering \tiny \obeylines
\begin{tcolorbox}[width=0.97\textwidth]
\begin{Verbatim}[breaklines, breaksymbolleft=]
"spatial_reasoning": {
    "short_description": "Visualizing and manipulating objects in space.",
    "long_description": "Spatial reasoning entails visualizing and manipulating objects in space. It involves understanding the relationships between different objects, such as their position, orientation, and movement relative to each other. Spatial reasoning is fundamental in fields like engineering, architecture, geography, and various forms of visual arts, enabling individuals to solve problems related to the physical arrangement and movement of object.",
    "abstract_example": "While arranging furniture, you visualize Object S_1 and Object S_2 to determine their optimal placement within the room.",
    "concrete_example": "A architect determining the best location for a window by visualizing the window and the surrounding walls to determine the optimal angle and height."
},
"social_reasoning": {
    "short_description": "Understanding social interactions and norms.",
    "long_description": "Social reasoning involves understanding social interactions and norms. It encompasses the ability to analyze and interpret social situations, recognize appropriate and inappropriate behaviors, and predict others' intentions, emotions, and thoughts. Effective social reasoning is crucial for building successful interpersonal relationships and navigating complex social environments.",
    "abstract_example": "Noticing that Person A behaves a certain way in Situation S, you adjust your own behavior (Behavior B) to interact effectively.",
    "concrete_example": "You notice that your friend looks upset after a conversation, so you decide to ask them if they are okay."
},
"moral_reasoning": {
    "short_description": "Deciding what is right or wrong based on ethical principles.",
    "long_description": "Moral reasoning is the process of deciding what is right or wrong based on ethical principles. It involves evaluating actions, intentions, and consequences to make judgments about moral issues. Moral reasoning is central to ethical decision-making and is influenced by various factors, including societal norms, personal values, and philosophical theories.",
    "abstract_example": "Considering that Action M could harm Person C, you decide it is morally wrong and choose an alternative that respects ethical principles.",
    "concrete_example": "Seeing someone drop their wallet, you decide to return it instead of keeping the money inside because it is the right thing to do."
}}
\end{Verbatim}
\end{tcolorbox}
\caption{\small Reasoning skill details.}
\label{fig:app_reason_skill}
\end{figure*}

\begin{figure*}[]
\captionsetup{skip=2pt}
\centering \tiny \obeylines
\begin{tcolorbox}[width=0.97\textwidth]
\begin{Verbatim}[breaklines, breaksymbolleft=]
### Commonsense-ness
{
    "task": "Evaluate the 'Commonsense-ness' of a multiple-choice commonsense question.",
    "evaluation_criteria": "Does answering the question rely solely on commonsense knowledge accessible to the general population, or does it require formal reasoning and specialized expertise beyond everyday understanding?",
    "rubric": {
        "1": "The question requires formal reasoning and specialized expertise to answer correctly. It demands advanced knowledge in a specific field, technical terminology, or in-depth understanding that goes beyond general life experience. The average person, relying only on commonsense knowledge, would find it challenging or impossible to select the correct answer without additional study or expertise.",
        "2": "The question can be addressed with some commonsense reasoning but may also require moderate specific knowledge or logical deduction. While not entirely dependent on formal expertise, it involves concepts or facts that are not universally known but can be reasoned through by an informed individual. The average person might answer correctly with thoughtful consideration but could also be misled without careful analysis.",
        "3": "The question is answerable using basic commonsense knowledge that is widely shared and understood by the general population. It does not rely on any specialized information or formal reasoning processes. The correct answer should be apparent to most people through everyday experience and general understanding of the world."
    }
}

### Complexity 
{
    "task": "Evaluate the 'Hardness/Complexity' of a commonsense question.",
    "evaluation_criteria": "How difficult is the question to understand and answer? Does it require minimal reasoning or a complex, multi-step thought process to identify the correct answer?",
    "rubric": {
        "1": "The question is very easy to understand, and the correct answer can be quickly identified with a single, straightforward reasoning step. It requires minimal cognitive effort, and most individuals can arrive at the correct answer almost immediately without confusion.",
        "2": "The question is relatively easy to understand, requiring only a couple of straightforward reasoning steps to identify the correct answer. While the question may introduce one or two elements that require brief consideration, the overall context remains clear. Most people can find the correct answer with a small amount of thought.",
        "3": "The question is moderately challenging, necessitating several reasoning steps to accurately comprehend and resolve. It introduces multiple elements or scenarios that require a careful thought process to integrate and analyze. Many individuals will need to pause and deliberately work through the connections or implications before reaching the correct answer.",
        "4": "The question is hard to comprehend and necessitates a complex thought process with multiple reasoning steps. It may involve abstract concepts, less obvious relationships, or misleading information that requires careful analysis. Individuals must invest significant cognitive effort to work through the complexities and identify the correct answer.",
        "5": "The question is very hard to comprehend and requires a long reasoning process with multiple reasoning steps to find the right answer. It demands high-level critical thinking, problem-solving skills, and possibly specialized knowledge. Only with thorough analysis and persistence can individuals navigate the complexity to arrive at the correct answer."
    }
}

### Expandability
{
    "task": "Evaluate the 'Expandability' of a commonsense question.",
    "evaluation_criteria": "To what extent can the question be expanded or elaborated upon to introduce additional complexity or dimensions?",
    "rubric": {
        "1": "The question cannot be expanded. It is inherently simplistic and covers a very narrow topic or scenario. There is little to no room for introducing additional elements, dimensions, or complexity without altering the fundamental nature of the question. The question stands effectively as a self-contained unit with minimal potential for elaboration.",
        "2": "The question has some potential for expansion. While it currently covers its intended scope adequately, there is moderate room to add a few additional elements or explore related themes that could introduce more complexity. The question can be expanded moderately by incorporating extra conditions, perspectives, or related scenarios, but such additions are not numerous.",
        "3": "The question can be significantly expanded to become a more complex question. It has ample scope for adding new dimensions, scenarios, or layers of reasoning. By introducing additional variables, conditional information, or intricate details, the question can transform into a more challenging problem that requires advanced reasoning and deeper comprehension."
    }
}
\end{Verbatim}
\end{tcolorbox}
\caption{\small Rubrics used for mCSQA Data Filtering Process}
\label{fig:app_rubric_mcsqa}
\end{figure*}

\begin{figure*}[]
\captionsetup{skip=2pt}
\centering \tiny \obeylines
\begin{tcolorbox}[width=0.97\textwidth]
\begin{Verbatim}[breaklines, breaksymbolleft=]
### Multicultureness
{
    "task": "evaluate the 'Multicultural-ness' of a commonsense cultural situation",
    "evaluation_criteria": "Does the situation involve interactions between multiple distinct cultures, reflecting a blend of practices, norms, or etiquette from each?",
    "rubric": {
        "1": "The situation is primarily rooted in a single culture, without significant influence or interaction from other cultural norms or practices. The interactions and behaviors exhibited are almost exclusively aligned with one cultural tradition, lacking a blend of cultural elements or considerations from another distinct culture.",
        "2": "The situation involves elements from two cultures, showing some level of cross-cultural interaction. While both cultural influences are present, the interaction may largely reflect the dominance of one culture over the other, with limited integration or blending of unique practices, norms, or etiquette from both cultures.",
        "3": "The situation reflects a rich blend of cultural interactions involving more than two distinct cultures. It demonstrates a balanced integration of diverse cultural practices, norms, or etiquette. The interactions and behaviors of the parties involved show a deep understanding and appreciation of multiple cultural perspectives, leading to an enriching multicultural exchange."
    }
}

### Commonsenseness
{
    "task": "evaluate the 'Commonsense-ness' of a cultural situation",
    "evaluation_criteria": "To what extent can the situation be understood and addressed using basic commonsense knowledge, without requiring specialized or expert reasoning?",
    "rubric": {
        "1": "The situation requires formal reasoning and specialized expertise to understand and address appropriately. It involves complex cultural nuances or specific knowledge that goes beyond general commonsense understanding. Responding effectively necessitates familiarity with detailed cultural protocols or insider knowledge.",
        "2": "The situation can be partially addressed using commonsense knowledge, but some elements require a deeper understanding or contextual insights that may not be readily apparent to someone without specific cultural awareness. While general reasoning can guide some actions, certain aspects benefit from additional cultural knowledge or experience.",
        "3": "The situation can be appropriately addressed using basic commonsense reasoning. It involves straightforward cultural interactions that do not demand specialized knowledge. Commonsense understanding of general social norms and human interactions is sufficient to respond suitably and effectively in this context."
        }
}

### Complexity
{
    "task": "Evaluate the 'Complexity' of a cultural situation.",
    "evaluation_criteria": "How intricate is the cultural situation in terms of nuances, number of cultural elements, perspectives, social dynamics, and interactions, requiring varying depths of understanding to navigate appropriately?",
    "rubric": {
        "1": "The situation is very simple, involving a single cultural aspect with straightforward practices and minimal perspectives or interactions. Understanding and responding require little to no specialized knowledge or awareness of cultural nuances.",
        "2": "The situation has minor complexity, incorporating a couple of cultural elements or perspectives with basic interactions. There are some cultural nuances, but they are easily understood with general awareness. Navigating the situation may require modest cultural sensitivity but is generally manageable.",
        "3": "The situation is moderately complex, involving several cultural elements, multiple perspectives, and noticeable social dynamics. Understanding and responding appropriately require some cultural knowledge and sensitivity to nuances. There is potential for misunderstandings without a moderate level of cultural competence.",
        "4": "The situation is complex, featuring numerous cultural elements, diverse perspectives, intricate social dynamics, and significant interactions. Navigating the situation effectively necessitates considerable cultural competence, an awareness of subtle nuances, and an understanding of how different cultural norms might conflict or interact.",
        "5": "The situation is highly complex, encompassing a multitude of deeply intertwined cultural elements, perspectives, and interactions. It includes profound cultural nuances, ambiguous social cues, and a high potential for misunderstandings. Expert knowledge and significant experience are required to address it appropriately, as the situation may involve conflicting norms and requires advanced cultural navigation skills."
    }
}

### Expandability
{
    "task": "Evaluate the 'Expandability' of a cultural situation",
    "evaluation_criteria": "Assess the potential for the situation to be expanded by including additional cultural dimensions, participants, interactions, and its adaptability to different contexts.",
    "rubric": {
        "1": "The situation is tightly defined within a single cultural framework, offering little room for the addition of new cultural dimensions. It does not easily support additional participants or interactions, requiring significant adaptation for expansion. It is context-specific and struggles to adapt to different settings or applications.",
        "2": "The situation allows for the inclusion of some additional cultural dimensions without drastically altering the core context. It can accommodate more participants or interactions with some adjustments to existing dynamics. There is some flexibility for adaptation to similar contexts or applications, albeit with moderate effort needed.",
        "3": "The situation is flexible and open, easily incorporating multiple new cultural dimensions or elements. It naturally supports additional participants and interactions without losing coherence. It is broadly applicable and adaptable across varied contexts and applications, maintaining core effectiveness and relevance."
    }
}
\end{Verbatim}
\end{tcolorbox}
\caption{\small Rubrics used for CultureBank Data Filtering Process}
\label{fig:app_rubric_culbank}
\end{figure*}

\begin{figure*}[]
\captionsetup{skip=2pt}
\centering \tiny \obeylines
\begin{tcolorbox}[width=0.99\textwidth]
\begin{Verbatim}[breaklines, breaksymbolleft=]
### LLM ROLE
You are a language model with advanced commonsense reasoning skills, capable of logical and analytical reasoning, heuristic and intuitive thinking, comparative and hypothetical analysis, and contextual and specialized understanding. 
    
### TASK DESCRIPTION
Given a multi-choice commonsense questions with the correct option, you task is to provide a "COMMONSENSE CONTEXT" to expand on the given question and a detailed "REASONING PROCESS" that involves multiple "REASONING STEPs" to arrive at the correct answer. 
+ A "COMMONSENSE CONTEXT" to the question refers to the background knowledge or additional details that are generally understood without requiring specialized knowledge, including factors such as time, place, social norms, cultural influences, and other relevant details that shape the understanding of the topic.
+ Each "REASONING STEP" should be an "ATOMIC REASONING STEP" — an Indivisible Unit of reasoning that predominantly utilizes one reasoning skill. It is a single, coherent thought process that cannot be broken down into smaller steps without losing its meaning. The "REASONING PROCESS" must be as efficent as possible, only using the minimum number of steps necessary, ensuring that each step is non-redundant and contributes to narrowing down the possible options by eliminating one or more answer choices.

### STEP-BY-STEP INSTRUCTIONS
Following these Step-by-Step Instructions:
1. Question Comprehension: Read the question carefully along with all the provided answer options.
2. Adding The "COMMONSENSE CONTEXT": Expand on the original question by providing an additional "COMMONSENSE CONTEXT". Ensure that the added context is relevant and enriches the understanding of the question.
3. Describe your Step-by-Step "REASONING PROCESS" to arrive at the correct answer. Each "ATOMIC REASONING STEP" must following this sequence:
    3.1. Choose a REASONING SKILL below to be used by the REASONING STEP:
        + inductive_reasoning: Drawing general conclusions from specific observations.
        + deductive_reasoning: Deriving specific conclusions from general premises.
        + abductive_reasoning: Forming hypotheses to explain observations.
        + analogical_reasoning: Drawing parallels between similar situations to infer conclusions.
        + counterfactual_reasoning: Considering alternative scenarios and outcomes that did not happen.
        + probabilistic_reasoning: Applying principles of probability to make inferences under uncertainty.
        + temporal_reasoning: Understanding sequences and durations of events.
        + spatial_reasoning: Visualizing and manipulating objects in space.
        + social_reasoning: Understanding social interactions and norms.
        + moral_reasoning: Deciding what is right or wrong based on ethical principles.
    3.2. Apply the choosen "REASONING SKILL": provide a concise explanation of how the chosen "REASONING SKILL" is applied to eliminate certain answer options or reinforce the correct answer option. Ensure the reasoning is clear and cannot be further divided into smaller steps.
    3.3. Eliminate Options: List the options eliminated in this step based on your reasoning.
    3.4. Update Possible Options: Provide the list of remaining possible options after this step.
4. Generate your output in the JSON format with the following structure:
    ```json
    {
      "commonsense_context": "context_text",
      "commonsense_question": "question_text",
      "options": {
          "A": "option_answer_text_A",
          ...
      },
      "correct_answer": ["answer_option", "answer_text"],
      "reasoning_process": {
          "reasoning_step_1": {
              "reasoning_skill": "reasoning_skill_name",
              "reasoning": "reasoning_text",
              "eliminated_options": [list_of_eliminated_options],
              "possible_options": [list_of_remaining_options]
          },
          ...
          "reasoning_step_n": {
              "reasoning_skill": "reasoning_skill_name",
              "reasoning": "reasoning_text",
              "eliminated_options": [list_of_eliminated_options],
              "possible_options": [list_of_remaining_options]
          }
      }
    }
    ```

### IN-CONTEXT EXAMPLE:
<.....>

### OUTPUT REMINDER
Ensure that your output follows the JSON structure as instructed and demonstrated in the in-context example.

### INPUT:
{
    "question": "What is the best way to experience a live performance?",
    "options": {
        "A": "watch play",
        "B": "go to theatre",
        "C": "open eyes",
        "D": "check showtimes",
        "E": "buy tickets"
    },
    "correct_answer": ["B", "go to theatre"]
}
\end{Verbatim}
\end{tcolorbox}
\caption{\small Prompt for Structured Reasoning Generation step for \textbf{mSCoRe-G} (English).}
\label{fig:app_prompt_reason_gen_general}
\end{figure*}

\begin{figure*}[]
\captionsetup{skip=2pt}
\centering \tiny \obeylines
\begin{tcolorbox}[width=0.97\textwidth]
\begin{Verbatim}[breaklines, breaksymbolleft=]
### LLM ROLE
You are a language model with advanced commonsense reasoning skills, capable of logical and analytical reasoning, heuristic and intuitive thinking, comparative and hypothetical analysis, and contextual and specialized understanding. 
       
### TASK DESCRIPTION
Given a multi-choice commonsense question with its options, your task is to modify and expand it to create a more complex question by expanding its context, modifying the question, adjusting the answer options, and adding an additional REASONING STEP. Your output should include the expanded context, the modified question, revised answer options, the correct answer, and a detailed "REASONING PROCESS".
       
### STEP-BY-STEP INSTRUCTIONS
Following these Step-by-Step Instructions:
1. Question Comprehension: Carefully read the given question and the context, and its answer options.
2. Context Expansion: adding additional backgound or situaltional details to the "COMMONSENSE CONTEXT" to add depth and reasoning requirements to the question.
3. Question Modificatioin:  Utilize the "EXPANDED COMMONSENSE CONTEXT" to craft a more complex question while maintaining its core concept and commonsense.
4. Option Adjustments: 
    + Adjust the existing answer options to align with the new complex question
    + Ensure the correct answer option remains semantically similar to the original
    + Introduce an additional plausible but incorrect option to increase the complexity of the question
    + Keep all answer options as concise as the originals
5. Reasoning Refinements: Refine the original "REASONING PROCESS" to fit the new context. The additional "ATOMIC REASONING STEP" must use one of the following "REASONING SKILLs":
    + inductive_reasoning: Drawing general conclusions from specific observations.
    + deductive_reasoning: Deriving specific conclusions from general premises.
    + abductive_reasoning: Forming hypotheses to explain observations.
    + analogical_reasoning: Drawing parallels between similar situations to infer conclusions.
    + counterfactual_reasoning: Considering alternative scenarios and outcomes that did not happen.
    + probabilistic_reasoning: Applying principles of probability to make inferences under uncertainty.
    + temporal_reasoning: Understanding sequences and durations of events.
    + spatial_reasoning: Visualizing and manipulating objects in space.
    + social_reasoning: Understanding social interactions and norms.
    + moral_reasoning: Deciding what is right or wrong based on ethical principles.
6. Format the Output using JSON format with the following structure:
    ```json
    {
      "commonsense_context": "context_text",
      "commonsense_question": "question_text",
      "options": {
          "A": "option_answer_text_A",
          ...
      },
      "correct_answer": ["answer_option", "answer_text"],
      "reasoning_process": {
          "reasoning_step_1": {
              "reasoning_skill": "reasoning_skill_name",
              "reasoning": "reasoning_text",
              "eliminated_options": [list_of_eliminated_options],
              "possible_options": [list_of_remaining_options]
          },
          ...
          "reasoning_step_n": {
              "reasoning_skill": "reasoning_skill_name",
              "reasoning": "reasoning_text",
              "eliminated_options": [list_of_eliminated_options],
              "possible_options": [list_of_remaining_options]
          }
      }
    }
    ```

### IN-CONTEXT EXAMPLE:
<.....>

### OUTPUT REMINDER
Ensure that your output follows the JSON structure as instructed and demonstrated in the in-context example.

### INPUT:
<.....>
\end{Verbatim}
\end{tcolorbox}
\caption{\small Prompt for Complexity Expansion step for \textbf{mSCoRe-G} (English).}
\label{fig:app_prompt_complex_exp}
\end{figure*}

\begin{figure*}[]
\captionsetup{skip=2pt}
\centering \tiny \obeylines
\begin{tcolorbox}[width=0.97\textwidth]
\begin{Verbatim}[breaklines, breaksymbolleft=]
### LLM ROLE
You are a language model with advanced commonsense reasoning skills, capable of logical and analytical reasoning, heuristic and intuitive thinking, comparative and hypothetical analysis, and contextual and specialized understanding. 
       

### TASK DESCRIPTION
Your task is to perform "Commonsense Implicitation," which involves combining a given "commonsense_context" with a "question" to generate a new, concise commonsense question that implicitly incorporates the original context. This process aims to evaluate the commonsense reasoning abilities of LLMs by ensuring that the implicit context preserves the original reasoning process and maintains the correctness of the answer.

### STEP-BY-STEP INSTRUCTIONS
Following these Step-by-Step Instructions:
1. Analyze the provided "commonsense_context" to understand the underlying assumptions and implicit knowledge required for reasoning
2. Examine the "commonsense_question" and its associated "options" to identify key elements essential for answering the question
3. Rewrite the "commonsense_question" by combining the original context and question to create a more new "commonsense_question" with an "IMPLICITLY IMPLIED COMMONSENSE CONTEXT". Ensure that the new question remains clear and understandable
4. Verify that the "REASONING PROCESS" remains unchanged in the transformed question, and confirm that the correct answer remains the same as in the original
5. Ensure that all answer options are reasonable, relevant, and maintain their original intent in the context of the rewritten question
6. Retain the structure and content of the "reasoning" section to reflect the logical steps supporting the correct answer. The  "ATOMIC REASONING STEP" must use one of the following "REASONING SKILLs":
    + inductive_reasoning: Drawing general conclusions from specific observations.
    + deductive_reasoning: Deriving specific conclusions from general premises.
    + abductive_reasoning: Forming hypotheses to explain observations.
    + analogical_reasoning: Drawing parallels between similar situations to infer conclusions.
    + counterfactual_reasoning: Considering alternative scenarios and outcomes that did not happen.
    + probabilistic_reasoning: Applying principles of probability to make inferences under uncertainty.
    + temporal_reasoning: Understanding sequences and durations of events.
    + spatial_reasoning: Visualizing and manipulating objects in space.
    + social_reasoning: Understanding social interactions and norms.
    + moral_reasoning: Deciding what is right or wrong based on ethical principles.

### IN-CONTEXT EXAMPLE:
<.....>

### OUTPUT REMINDER
Ensure that your output follows the JSON structure as instructed and demonstrated in the in-context example.

### INPUT:
<.....>
\end{Verbatim}
\end{tcolorbox}
\caption{\small Prompt for Commonsense Implicitation step for \textbf{mSCoRe-G} (English).}
\label{fig:app_prompt_com_impl}
\end{figure*}

\begin{figure*}[]
\captionsetup{skip=2pt}
\centering \tiny \obeylines
\begin{tcolorbox}[width=0.97\textwidth]
\begin{Verbatim}[breaklines, breaksymbolleft=]
### ORIGINAL QUESTION FROM mCSQA:
{
    "question": "What is the best way to experience a live performance?",
    "options": {
        "A": "watch play",
        "B": "go to theatre",
        "C": "open eyes",
        "D": "check showtimes",
        "E": "buy tickets"
    },
    "correct_answer": [
        "B",
        "go to theatre"
    ]
}

### COMPEXITY LEVEL 0:    
{
  "commonsense_context": "Experiencing a live performance, such as a play, concert, or other theatrical event, typically involves attending a venue where the performance is being held. The atmosphere, acoustics, and visual elements of a live performance are best appreciated in person, where the audience can fully engage with the performers and the production. This is a social and cultural activity that often involves planning, such as purchasing tickets and checking showtimes, but the core experience is being physically present at the venue.",
  "commonsense_question": "What is the best way to experience a live performance?",
  "options": {
      "A": "watch play",
      "B": "go to theatre",
      "C": "open eyes",
      "D": "check showtimes",
      "E": "buy tickets"
  },
  "correct_answer": ["B", "go to theatre"],
  "reasoning_process": {
      "reasoning_step_1": {
          "reasoning_skill": "deductive_reasoning",
          "reasoning": "To experience a live performance, one must be physically present where the performance is taking place. This means attending the venue, such as a theatre, where the live event is held.",
          "eliminated_options": ["A", "C"],
          "possible_options": ["B", "D", "E"]
      },
      "reasoning_step_2": {
          "reasoning_skill": "temporal_reasoning",
          "reasoning": "Checking showtimes and buying tickets are preparatory actions that occur before the actual experience of the live performance. They are necessary steps but do not constitute the experience itself.",
          "eliminated_options": ["D", "E"],
          "possible_options": ["B"]
      }
  }
}

### COMPLEXITY LEVEL 1
{
  "commonsense_context": "Experiencing a live performance, such as a play, concert, or other theatrical event, typically involves attending a venue where the performance is being held. The atmosphere, acoustics, and visual elements of a live performance are best appreciated in person, where the audience can fully engage with the performers and the production. This is a social and cultural activity that often involves planning, such as purchasing tickets and checking showtimes, but the core experience is being physically present at the venue. Additionally, live performances often include unique interactions between the audience and performers, such as applause, laughter, or even participation, which enhance the overall experience.",
  "commonsense_question": "In what way can you fully immerse yourself in the unique atmosphere and interactions of a live performance?",
  "options": {
      "A": "watch a recording at home",
      "B": "attend the theatre in person",
      "C": "read reviews online",
      "D": "check showtimes regularly",
      "E": "buy tickets in advance",
      "F": "listen to a podcast about the performance"
  },
  "correct_answer": ["B", "attend the theatre in person"],
  "reasoning_process": {
      "reasoning_step_1": {
          "reasoning_skill": "deductive_reasoning",
          "reasoning": "To fully immerse oneself in the unique atmosphere and interactions of a live performance, one must be physically present at the venue. This allows for direct engagement with the performance and the performers.",
          "eliminated_options": ["A", "C"],
          "possible_options": ["B", "D", "E","F"]
      },
\end{Verbatim}
\end{tcolorbox}
\end{figure*}

\begin{figure*}[]
\captionsetup{skip=2pt}
\centering \tiny \obeylines
\begin{tcolorbox}[width=0.97\textwidth]
\begin{Verbatim}[breaklines, breaksymbolleft=]
      "reasoning_step_2": {
          "reasoning_skill": "temporal_reasoning",
          "reasoning": "Checking showtimes and buying tickets are preparatory actions that occur before attending the performance. They are necessary for planning but do not provide the immersive experience itself.",
          "eliminated_options": ["D", "E"],
          "possible_options": ["B", "F"]
      },
      "reasoning_step_3": {
          "reasoning_skill": "social_reasoning",
          "reasoning": "Being present at the theatre allows for social interactions and shared experiences with other audience members and the performers, which are integral to the live performance experience.",
          "eliminated_options": ["F"],
          "possible_options": ["B"]
      }
  }
}

### COMPLEXITY LEVEL 2:
{
  "commonsense_context": "Experiencing a live performance, such as a play, concert, or other theatrical event, typically involves attending a venue where the performance is being held. The atmosphere, acoustics, and visual elements of a live performance are best appreciated in person, where the audience can fully engage with the performers and the production. This is a social and cultural activity that often involves planning, such as purchasing tickets and checking showtimes, but the core experience is being physically present at the venue. Additionally, live performances often include unique interactions between the audience and performers, such as applause, laughter, or even participation, which enhance the overall experience. Furthermore, attending a live performance can create lasting memories and a sense of community among attendees, as they share the emotional highs and lows of the performance together.",
  "commonsense_question": "How can you create lasting memories and fully immerse yourself in the unique atmosphere and interactions of a live performance, while also contributing to the communal experience?",
  "options": {
      "A": "watch a recording at home",
      "B": "attend the theatre in person",
      "C": "read reviews online",
      "D": "check showtimes regularly",
      "E": "buy tickets in advance",
      "F": "listen to a podcast about the performance",
      "G": "participate in a post-show discussion"
  },
  "correct_answer": ["B", "attend the theatre in person"],
  "reasoning_process": {
      "reasoning_step_1": {
          "reasoning_skill": "deductive_reasoning",
          "reasoning": "To create lasting memories and fully immerse oneself in the unique atmosphere and interactions of a live performance, one must be physically present at the venue. This allows for direct engagement with the performance and the performers.",
          "eliminated_options": ["A", "C", "F"],
          "possible_options": ["B", "D", "E", "G"]
      },
      "reasoning_step_2": {
          "reasoning_skill": "temporal_reasoning",
          "reasoning": "Checking showtimes and buying tickets are preparatory actions that occur before attending the performance. They are necessary for planning but do not provide the immersive experience itself.",
          "eliminated_options": ["D", "E"],
          "possible_options": ["B", "G"]
      },
      "reasoning_step_3": {
          "reasoning_skill": "social_reasoning",
          "reasoning": "Being present at the theatre allows for social interactions and shared experiences with other audience members and the performers, which are integral to the live performance experience. While participating in a post-show discussion can enhance the communal experience, it does not replace the immersive experience of attending the performance itself.",
          "eliminated_options": ["G"],
          "possible_options": ["B"]
      }
  }
}

### COMPEXITY LEVEL 3:
{
  "commonsense_context": "Experiencing a live performance, such as a play, concert, or other theatrical event, typically involves attending a venue where the performance is being held. The atmosphere, acoustics, and visual elements of a live performance are best appreciated in person, where the audience can fully engage with the performers and the production. This is a social and cultural activity that often involves planning, such as purchasing tickets and checking showtimes, but the core experience is being physically present at the venue. Additionally, live performances often include unique interactions between the audience and performers, such as applause, laughter, or even participation, which enhance the overall experience. Furthermore, attending a live performance can create lasting memories and a sense of community among attendees, as they share the emotional highs and lows of the performance together. In recent times, some performances have also incorporated digital elements, allowing for a hybrid experience where audiences can engage both in-person and online, adding a new dimension to the traditional live performance.",
  "commonsense_question": "In the context of a modern live performance that incorporates both in-person and digital elements, how can you create lasting memories and fully immerse yourself in the unique atmosphere and interactions, while also contributing to the communal experience?",
\end{Verbatim}
\end{tcolorbox}
\end{figure*}

\begin{figure*}[]
\captionsetup{skip=2pt}
\centering \tiny \obeylines
\begin{tcolorbox}[width=0.97\textwidth]
\begin{Verbatim}[breaklines, breaksymbolleft=]
  "options": {
      "A": "watch a recording at home",
      "B": "attend the theatre in person",
      "C": "read reviews online",
      "D": "check showtimes regularly",
      "E": "buy tickets in advance",
      "F": "listen to a podcast about the performance",
      "G": "participate in a post-show discussion",
      "H": "engage with digital elements during the performance"
  },
  "correct_answer": ["B", "attend the theatre in person"],
  "reasoning_process": {
      "reasoning_step_1": {
          "reasoning_skill": "deductive_reasoning",
          "reasoning": "To create lasting memories and fully immerse oneself in the unique atmosphere and interactions of a live performance, one must be physically present at the venue. This allows for direct engagement with the performance and the performers.",
          "eliminated_options": ["A", "C", "F"],
          "possible_options": ["B", "D", "E", "G", "H"]
      },
      "reasoning_step_2": {
          "reasoning_skill": "temporal_reasoning",
          "reasoning": "Checking showtimes and buying tickets are preparatory actions that occur before attending the performance. They are necessary for planning but do not provide the immersive experience itself.",
          "eliminated_options": ["D", "E"],
          "possible_options": ["B", "G", "H"]
      },
      "reasoning_step_3": {
          "reasoning_skill": "social_reasoning",
          "reasoning": "Being present at the theatre allows for social interactions and shared experiences with other audience members and the performers, which are integral to the live performance experience. While participating in a post-show discussion can enhance the communal experience, it does not replace the immersive experience of attending the performance itself.",
          "eliminated_options": ["G"],
          "possible_options": ["B", "H"]
      },
      "reasoning_step_4": {
          "reasoning_skill": "analogical_reasoning",
          "reasoning": "Engaging with digital elements during the performance can enhance the experience but is analogous to supplementary activities. The core immersive experience is still best achieved by being physically present.",
          "eliminated_options": ["H"],
          "possible_options": ["B"]
      }
  }
}
\end{Verbatim}
\end{tcolorbox}
\caption{\small An example from \textbf{mSCoRe-G} for complexity level 0 to 3 (English).}
\label{fig:app_general_example}
\end{figure*}

\begin{figure*}[]
\captionsetup{skip=2pt}
\centering \scriptsize \obeylines
\begin{tcolorbox}[width=0.9\textwidth]
\begin{Verbatim}[breaklines, breaksymbolleft=]
### LLM ROLE
You are a language model with advanced commonsense reasoning skills, capable of logical and analytical reasoning, heuristic and intuitive thinking, comparative and hypothetical analysis, and contextual and specialized understanding. 
       

### TASK DESCRIPTION
Your task is to create a multiple-choice commonsense question based on a given cultural situation in the following format:
{
  "cultural_topic": "culture group - topic - scenario",
  "social_context": "settings the behavior takes place",
  "actor": "who exhibit the behavior",
  "question": "the commonsense question regarding the actor's behavior",
  "actor_behavior": "behavior of the actor - which are highly agreed upon (the correct answer option)",
  "recipient": "recipient of the action",
  "relation": "relation between the actor and the recipient",
  "recipient_behavior": "behavior of the recipient",
}
The question should implicitly incorporate the cultural context, challenging the AI's ability to utilize commonsense reasoning to arrive at the correct answer. The goal is to test and enhance the AI's understanding of cultural norms and behaviors in a specific setting.
Provide the detailed "REASONING PROCESS" the arrive at the correct anwser option that involves multiple "REASONING STEPs" to arrive at the correct answer. Each "REASONING STEP" should be an "ATOMIC REASONING STEP" — an Indivisible Unit of reasoning that predominantly utilizes one reasoning skill. It is a single, coherent thought process that cannot be broken down into smaller steps without losing its meaning. The "REASONING PROCESS" must be as efficent as possible, only using the minimum number of steps necessary, ensuring that each step is non-redundant and contributes to narrowing down the possible options by eliminating one or more answer choices.
       

### STEP-BY-STEP INSTRUCTIONS
Following these Step-by-Step Instructions:
1. Analyze the Provided Cultural Situation: Review the details of the cultural group, context, actor behaviors, and other descriptions to understand the key elements of the situation.
2. Adding The "COMMONSENSE CONTEXT": Based on the context given in the input, A "COMMONSENSE CONTEXT" to the question refers to the background knowledge or additional details that are generally understood without requiring specialized knowledge, including factors such as time, place, social norms, cultural influences, and other relevant details that shape the understanding of the topic.
3. Create the "Commonsense Question": Combine the cultural context and the persona's inquiry to formulate a concise question. Ensure the question IMPLICITLY incorporates the original context without explicitly stating it. Create the correct answer option based on the "actor_behavior"
4. Provide Other Answer Options: Create 5 multiple-choice options (including the correct answer from the previous step). Two of which should be plausible options. The other two should be distractors that are relevant and reasonable but incorrect based on the cultural context.
5. Describe your Step-by-Step "REASONING PROCESS" to arrive at the correct answer. Each "ATOMIC REASONING STEP" must following this sequence:
    5.1. Choose a "REASONING SKILL" below to be used by the "REASONING STEP":
        + inductive_reasoning: Drawing general conclusions from specific observations.
        + deductive_reasoning: Deriving specific conclusions from general premises.
        + abductive_reasoning: Forming hypotheses to explain observations.
        + analogical_reasoning: Drawing parallels between similar situations to infer conclusions.
        + counterfactual_reasoning: Considering alternative scenarios and outcomes that did not happen.
        + probabilistic_reasoning: Applying principles of probability to make inferences under uncertainty.
        + temporal_reasoning: Understanding sequences and durations of events.
        + spatial_reasoning: Visualizing and manipulating objects in space.
        + social_reasoning: Understanding social interactions and norms.
        + moral_reasoning: Deciding what is right or wrong based on ethical principles.
    5.2. Apply the choosen "REASONING SKILL": provide a concise explanation of how the chosen "REASONING SKILL" is applied to eliminate certain answer options or reinforce the correct answer option. Ensure the reasoning is clear and cannot be further divided into smaller steps.  
    5.3. Eliminate Options: List the options eliminated in this step based on your reasoning.
    5.4. Update Possible Options: Provide the list of remaining possible options after this step.
\end{Verbatim}
\end{tcolorbox}
\end{figure*}

\begin{figure*}[]
\captionsetup{skip=2pt}
\centering \scriptsize \obeylines
\begin{tcolorbox}[width=0.9\textwidth]
\begin{Verbatim}[breaklines, breaksymbolleft=]    
6. Generate your output in the JSON format with the following structure:

    ```json
    {
      "commonsense_context": "context_text",
      "commonsense_question": "question_text",
      "options": {
          "A": "option_answer_text_A",
          ...
      },
      "correct_answer": ["answer_option", "answer_text"],
      "reasoning_process": {
          "reasoning_step_1": {
              "reasoning_skill": "reasoning_skill_name",
              "reasoning": "reasoning_text",
              "eliminated_options": [list_of_eliminated_options],
              "possible_options": [list_of_remaining_options]
          },
          ...
          "reasoning_step_n": {
              "reasoning_skill": "reasoning_skill_name",
              "reasoning": "reasoning_text",
              "eliminated_options": [list_of_eliminated_options],
              "possible_options": [list_of_remaining_options]
          }
      }
    }
    ```

### IN-CONTEXT EXAMPLE:
<.....>

### OUTPUT REMINDER
Ensure that your output follows the JSON structure as instructed and demonstrated in the in-context example.

### INPUT:
{
    "cultural_topic": "American culture - Dress Codes - Travel Advising",
    "social_context": "In public settings within American culture, it is common for people to dress casually, often opting for comfortable clothing such as sweatpants while still adhering to dress codes. This relaxed approach to attire is widely regarded as the norm by a significant portion of the sampled population. It reflects a preference for comfort and practicality in daily dress, showcasing a relaxed and informal attitude towards clothing choices in various public settings.",
    "actor": "people - A business professional from a formal corporate background, planning a first-time trip to the United States for a business conference, eager to blend in and avoid any potential faux pas",
    "question": "I'm gearing up for a big conference in the States and I'm a bit worried about what to wear. I come from a formal work environment and I don't want to stand out in a negative way. Can you give me some tips on what kind of attire would be appropriate for a business setting over there? Should I be concerned about anything specific?",
    "actor_behavior": "dress casually, often in comfortable clothing, with a preference for sweatpants and following dress codes",
    "recipient": "None",
    "relation": "None",
    "recipient_behavior": "None"
}
\end{Verbatim}
\end{tcolorbox}
\caption{\small Structured Reasoning Generation Prompt for \textbf{mSCoRe-S}.}
\label{fig:app_prompt_reason_gen_social}
\end{figure*}

\begin{figure*}[]
\captionsetup{skip=2pt}
\centering \scriptsize \obeylines
\begin{tcolorbox}[width=0.9\textwidth]
\begin{Verbatim}[breaklines, breaksymbolleft=] 
### ORIGINAL INSTANCE FROM CULTUREBANK:
{
    "cultural_topic": "Germans culture - Education and Technology - Travel Advising",
    "social_context": "In German schools, both the educational institutions and students actively participate in compulsory swimming education, which includes separate classes for students with limited swimming skills. The goal of this initiative is to teach swimming skills and promote integration, with students, including those with immigrant parents, participating in swimming lessons. It is noteworthy that Muslim students are accommodated by wearing burqinis during these swimming classes, showcasing inclusivity and respect for diverse cultural practices. This swimming education typically takes place in 5th or 6th grade and lasts for 2 years, resulting in students acquiring advanced swimming skills. While not universally embraced, a significant portion of the sampled population considers this practice as a standard part of the educational experience in German schools.",
    "actor": "German schools and students - A concerned parent planning a family trip to Germany, looking for a comprehensive and inclusive educational experience for their children",
    "question": "I'm planning a family trip to Germany and I want my kids to make the most of their time there. I've heard that schools there offer some unique educational experiences. I'm particularly interested in finding activities that are both fun and educational for my kids, especially ones that can help them learn new skills. Could you recommend some family-friendly programs that might be a good fit for us, keeping in mind that we have a diverse family background?",
    "actor_behavior": "provide and attend compulsory swimming education, including separate classes for those with limited swimming skills",
    "recipient": "German students, including those with immigrant parents",
    "relation": "educational institution and attendees",
    "recipient_behavior": "participate in swimming lessons, including wearing burqinis for Muslim students"
}


### COMPEXITY LEVEL 0:
{
  "commonsense_context": "In Germany, swimming education is an integral part of the school curriculum, aimed at teaching essential swimming skills and promoting inclusivity. This program is designed to accommodate students from diverse backgrounds, including those with immigrant parents and Muslim students, who are allowed to wear burqinis. The initiative is generally well-received and considered a standard educational practice, providing students with valuable life skills and fostering integration.",
  "commonsense_question": "What unique educational experience in German schools could be particularly beneficial for a diverse family visiting Germany, looking for fun and educational activities for their children?",
  "options": {
      "A": "Participate in compulsory swimming education that includes accommodations for diverse cultural practices.",
      "B": "Enroll in advanced mathematics classes to enhance analytical skills.",
      "C": "Join a German language immersion program to improve language proficiency.",
      "D": "Attend a local history tour to learn about German culture and heritage.",
      "E": "Take part in a cooking class to explore traditional German cuisine."
  },
  "correct_answer": ["A", "Participate in compulsory swimming education that includes accommodations for diverse cultural practices."],
  "reasoning_process": {
      "reasoning_step_1": {
          "reasoning_skill": "social_reasoning",
          "reasoning": "Understanding that the swimming education program in German schools is designed to be inclusive and accommodate diverse cultural backgrounds, making it a unique and suitable experience for a diverse family.",
          "eliminated_options": ["B", "C"],
          "possible_options": ["A", "D", "E"]
      },
      "reasoning_step_2": {
          "reasoning_skill": "deductive_reasoning",
          "reasoning": "Considering the context of the question, which emphasizes fun and educational activities, swimming education stands out as it combines physical activity with skill acquisition, unlike a history tour which is more passive.",
          "eliminated_options": ["D", "E"],
          "possible_options": ["A"]
      }
  }
}
\end{Verbatim}
\end{tcolorbox}
\end{figure*}

\begin{figure*}[]
\captionsetup{skip=2pt}
\centering \scriptsize \obeylines
\begin{tcolorbox}[width=0.9\textwidth]
\begin{Verbatim}[breaklines, breaksymbolleft=] 
### COMPLEXITY LEVEL 1:
{
  "commonsense_context": "In Germany, swimming education is an integral part of the school curriculum, aimed at teaching essential swimming skills and promoting inclusivity. This program is designed to accommodate students from diverse backgrounds, including those with immigrant parents and Muslim students, who are allowed to wear burqinis. The initiative is generally well-received and considered a standard educational practice, providing students with valuable life skills and fostering integration. Additionally, German schools often collaborate with local community centers to offer these swimming lessons, ensuring that students have access to safe and well-maintained facilities. This collaboration also allows for the inclusion of parents in some sessions, promoting family involvement in the educational process.",
  "commonsense_question": "For a diverse family visiting Germany, interested in engaging in both educational and community activities, what unique experience offered by German schools could be particularly beneficial?",
  "options": {
    "A": "Participate in compulsory swimming education that includes accommodations for diverse cultural practices and involves community engagement.",
    "B": "Enroll in advanced mathematics classes to enhance analytical skills.",
    "C": "Join a German language immersion program to improve language proficiency.",
    "D": "Attend a local history tour to learn about German culture and heritage.",
    "E": "Take part in a cooking class to explore traditional German cuisine.",
    "F": "Engage in a community art project to express cultural diversity."
  },
  "correct_answer": ["A", "Participate in compulsory swimming education that includes accommodations for diverse cultural practices and involves community engagement."],
  "reasoning_process": {
    "reasoning_step_1": {
      "reasoning_skill": "social_reasoning",
      "reasoning": "Understanding that the swimming education program in German schools is designed to be inclusive and accommodate diverse cultural backgrounds, making it a unique and suitable experience for a diverse family. The program's collaboration with community centers further enhances its appeal by involving the family in the local community.",
      "eliminated_options": ["B", "C"],
      "possible_options": ["A", "D", "E", "F"]
    },
    "reasoning_step_2": {
      "reasoning_skill": "deductive_reasoning",
      "reasoning": "Considering the context of the question, which emphasizes educational and community activities, swimming education stands out as it combines physical activity, skill acquisition, and community involvement, unlike a history tour which is more passive.",
      "eliminated_options": ["D"],
      "possible_options": ["A", "E", "F"]
    },
    "reasoning_step_3": {
      "reasoning_skill": "abductive_reasoning",
      "reasoning": "While a cooking class and a community art project can be educational and fun, they do not offer the same level of inclusivity, skill development, and structured community engagement as the swimming program, which is a part of the school curriculum.",
      "eliminated_options": ["E", "F"],
      "possible_options": ["A"]
    }
  }
}
\end{Verbatim}
\end{tcolorbox}
\end{figure*}

\begin{figure*}[]
\captionsetup{skip=2pt}
\centering \scriptsize \obeylines
\begin{tcolorbox}[width=0.9\textwidth]
\begin{Verbatim}[breaklines, breaksymbolleft=] 
### COMPLEXITY LEVEL 2:
{
  "commonsense_context": "In Germany, swimming education is an integral part of the school curriculum, aimed at teaching essential swimming skills and promoting inclusivity. This program is designed to accommodate students from diverse backgrounds, including those with immigrant parents and Muslim students, who are allowed to wear burqinis. The initiative is generally well-received and considered a standard educational practice, providing students with valuable life skills and fostering integration. Additionally, German schools often collaborate with local community centers to offer these swimming lessons, ensuring that students have access to safe and well-maintained facilities. This collaboration also allows for the inclusion of parents in some sessions, promoting family involvement in the educational process. Furthermore, these swimming programs often include cultural exchange activities, where students and their families can share and learn about each other's traditions, enhancing mutual understanding and respect.",
  "commonsense_question": "For a diverse family visiting Germany, interested in engaging in both educational and community activities, what unique experience offered by German schools could be particularly beneficial, especially in terms of cultural exchange and inclusivity?",
  "options": {
    "A": "Participate in compulsory swimming education that includes accommodations for diverse cultural practices, involves community engagement, and offers cultural exchange opportunities.",
    "B": "Enroll in advanced mathematics classes to enhance analytical skills.",
    "C": "Join a German language immersion program to improve language proficiency.",
    "D": "Attend a local history tour to learn about German culture and heritage.",
    "E": "Take part in a cooking class to explore traditional German cuisine.",
    "F": "Engage in a community art project to express cultural diversity.",
    "G": "Participate in a multicultural festival organized by the school."
  },
  "correct_answer": ["A", "Participate in compulsory swimming education that includes accommodations for diverse cultural practices, involves community engagement, and offers cultural exchange opportunities."],
  "reasoning_process": {
    "reasoning_step_1": {
      "reasoning_skill": "social_reasoning",
      "reasoning": "Understanding that the swimming education program in German schools is designed to be inclusive and accommodate diverse cultural backgrounds, making it a unique and suitable experience for a diverse family. The program's collaboration with community centers further enhances its appeal by involving the family in the local community.",
      "eliminated_options": ["B", "C"],
      "possible_options": ["A", "D", "E", "F", "G"]
    },
    "reasoning_step_2": {
      "reasoning_skill": "deductive_reasoning",
      "reasoning": "Considering the context of the question, which emphasizes educational and community activities, swimming education stands out as it combines physical activity, skill acquisition, and community involvement, unlike a history tour which is more passive.",
      "eliminated_options": ["D"],
      "possible_options": ["A", "E", "F", "G"]
    },
    "reasoning_step_3": {
      "reasoning_skill": "abductive_reasoning",
      "reasoning": "While a cooking class and a community art project can be educational and fun, they do not offer the same level of inclusivity, skill development, and structured community engagement as the swimming program, which is a part of the school curriculum.",
      "eliminated_options": ["E", "F"],
      "possible_options": ["A", "G"]
    },
    "reasoning_step_4": {
      "reasoning_skill": "analogical_reasoning",
      "reasoning": "Comparing the swimming program with the multicultural festival, the swimming program offers a more structured and ongoing opportunity for cultural exchange and skill development, whereas the festival is a one-time event.",
      "eliminated_options": ["G"],
      "possible_options": ["A"]
    }
  }
}
\end{Verbatim}
\end{tcolorbox}
\end{figure*}

\begin{figure*}[]
\captionsetup{skip=2pt}
\centering \scriptsize \obeylines
\begin{tcolorbox}[width=0.9\textwidth]
\begin{Verbatim}[breaklines, breaksymbolleft=] 
### COMPLEXITY LEVEL 3:
{
  "commonsense_context": "In Germany, swimming education is an integral part of the school curriculum, aimed at teaching essential swimming skills and promoting inclusivity. This program is designed to accommodate students from diverse backgrounds, including those with immigrant parents and Muslim students, who are allowed to wear burqinis. The initiative is generally well-received and considered a standard educational practice, providing students with valuable life skills and fostering integration. Additionally, German schools often collaborate with local community centers to offer these swimming lessons, ensuring that students have access to safe and well-maintained facilities. This collaboration also allows for the inclusion of parents in some sessions, promoting family involvement in the educational process. Furthermore, these swimming programs often include cultural exchange activities, where students and their families can share and learn about each other's traditions, enhancing mutual understanding and respect. The program also emphasizes water safety, which is a crucial skill for everyone, and includes sessions on the importance of respecting different cultural practices in shared spaces.",
  "commonsense_question": "For a diverse family visiting Germany, interested in engaging in both educational and community activities, what unique experience offered by German schools could be particularly beneficial, especially in terms of cultural exchange, inclusivity, and learning essential life skills like water safety?",
  "options": {
    "A": "Participate in compulsory swimming education that includes accommodations for diverse cultural practices, involves community engagement, and offers cultural exchange opportunities.",
    "B": "Enroll in advanced mathematics classes to enhance analytical skills.",
    "C": "Join a German language immersion program to improve language proficiency.",
    "D": "Attend a local history tour to learn about German culture and heritage.",
    "E": "Take part in a cooking class to explore traditional German cuisine.",
    "F": "Engage in a community art project to express cultural diversity.",
    "G": "Participate in a multicultural festival organized by the school.",
    "H": "Join a water safety workshop that includes cultural sensitivity training."
  },
  "correct_answer": ["A", "Participate in compulsory swimming education that includes accommodations for diverse cultural practices, involves community engagement, and offers cultural exchange opportunities."],
  "reasoning_process": {
    "reasoning_step_1": {
      "reasoning_skill": "social_reasoning",
      "reasoning": "Understanding that the swimming education program in German schools is designed to be inclusive and accommodate diverse cultural backgrounds, making it a unique and suitable experience for a diverse family. The program's collaboration with community centers further enhances its appeal by involving the family in the local community.",
      "eliminated_options": ["B", "C"],
      "possible_options": ["A", "D", "E", "F", "G", "H"]
    },
    "reasoning_step_2": {
      "reasoning_skill": "deductive_reasoning",
      "reasoning": "Considering the context of the question, which emphasizes educational and community activities, swimming education stands out as it combines physical activity, skill acquisition, and community involvement, unlike a history tour which is more passive.",
      "eliminated_options": ["D"],
      "possible_options": ["A", "E", "F", "G", "H"]
    },
    "reasoning_step_3": {
      "reasoning_skill": "abductive_reasoning",
      "reasoning": "While a cooking class and a community art project can be educational and fun, they do not offer the same level of inclusivity, skill development, and structured community engagement as the swimming program, which is a part of the school curriculum.",
      "eliminated_options": ["E", "F"],
      "possible_options": ["A", "G", "H"]
    },
\end{Verbatim}
\end{tcolorbox}
\end{figure*}

\begin{figure*}[]
\captionsetup{skip=2pt}
\centering \scriptsize \obeylines
\begin{tcolorbox}[width=0.9\textwidth]
\begin{Verbatim}[breaklines, breaksymbolleft=] 
    "reasoning_step_4": {
      "reasoning_skill": "analogical_reasoning",
      "reasoning": "Comparing the swimming program with the multicultural festival, the swimming program offers a more structured and ongoing opportunity for cultural exchange and skill development, whereas the festival is a one-time event.",
      "eliminated_options": ["G"],
      "possible_options": ["A", "H"]
    },
    "reasoning_step_5": {
      "reasoning_skill": "probabilistic_reasoning",
      "reasoning": "While a water safety workshop with cultural sensitivity training is beneficial, the swimming program is more comprehensive, offering ongoing lessons that include water safety as part of a broader curriculum.",
      "eliminated_options": ["H"],
      "possible_options": ["A"]
    }
  }
}

\end{Verbatim}
\end{tcolorbox}
\caption{\small An example from \textbf{mSCoRe-S} for complexity level 0 to 3 (English).}
\label{fig:app_social_example}
\end{figure*}









\end{document}